\PassOptionsToPackage{table,dvipsnames}{xcolor}

\documentclass{article}
\usepackage{iclr2025_conference,times}

\iclrfinalcopy 

\usepackage{bold-extra}
\usepackage{times}
\usepackage{latexsym}
\usepackage{inconsolata}
\usepackage[most]{tcolorbox}
\usepackage{soul}
\usepackage{amsmath,amsfonts,amssymb,amsthm}
\usepackage[export]{adjustbox}
\usepackage{wrapfig}
\usepackage{soul}
\usepackage{makecell}
\usepackage{multirow}
\usepackage{balance}
\usepackage{xspace}
\usepackage[english]{babel}
\usepackage{bm}
\usepackage{xurl}
\usepackage{algorithm,algorithmic}
\usepackage{arydshln} 
\usepackage{enumitem}
\usepackage{circledsteps}
\usepackage{subfigure}

\usepackage[utf8]{inputenc} 
\usepackage[T1]{fontenc}    
\usepackage[
bookmarksopen,
bookmarksdepth=2,
breaklinks=true,
colorlinks=true,
linkcolor=linkcolour,
citecolor=citecolour,
filecolor=urlcolour,
urlcolor=urlcolour,
backref=page]{hyperref}
\usepackage{url}            
\usepackage{booktabs}       
\usepackage{amsfonts}       
\usepackage{nicefrac}       
\usepackage{microtype}      
\usepackage{lipsum}		
\usepackage{graphicx}
\usepackage{natbib}
\usepackage{doi}

\usepackage{titletoc}

\definecolor{linkcolour}{rgb}{0,0.2,0.6}
\definecolor{urlcolour}{rgb}{0,0.6,0.2}
\definecolor{citecolour} {rgb}{0.8,0,0.8}

\newcommand{\method}{SRFT\xspace}

\usepackage{listings}
\title{\method: A Single-Stage Method with Supervised and Reinforcement Fine-Tuning for Reasoning}

\author{Authors}

\author{\textbf{Yuqian Fu$^\spadesuit$$^\clubsuit$\thanks{Equal contribution. $^\dag$Corresponding authors. $^\ddag$Work in progress.}}~~,
\textbf{Tinghong Chen$^\spadesuit$$^\clubsuit$$^{*}$},
\textbf{Jiajun Chai$^\diamondsuit$},
\textbf{Xihuai Wang$^\heartsuit$},
\textbf{Songjun Tu$^\spadesuit$$^\clubsuit$},\\
\textbf{Guojun Yin$^\diamondsuit$},
\textbf{Wei Lin$^\diamondsuit$},
\textbf{Qichao Zhang$^\spadesuit$$^\clubsuit$},
\textbf{Yuanheng Zhu$^\spadesuit$$^\clubsuit$$^\dag$},
\textbf{Dongbin Zhao$^\spadesuit$$^\clubsuit$$^\dag$}
\\
 $^\spadesuit$~Institute of Automation, Chinese Academy of Sciences\\
 $^\clubsuit$~School of Artificial Intelligence, University of Chinese Academy of Sciences \\
 $^\diamondsuit$~Meituan \quad
 $^\heartsuit$~Shanghai Jiao Tong University
\\
\texttt{\{fuyuqian2022,yuanheng.zhu\}@ia.ac.cn}
}

\begin{document}

\maketitle

\vspace{-1em}
\begin{abstract}
Large language models (LLMs) have achieved remarkable progress in reasoning tasks, yet the optimal integration of Supervised Fine-Tuning (SFT) and Reinforcement Learning (RL) remains a fundamental challenge. Through comprehensive analysis of token distributions, learning dynamics, and integration mechanisms from entropy-based perspectives, we reveal key differences between these paradigms: SFT induces coarse-grained global changes to LLM policy distributions, while RL performs fine-grained selective optimizations, with entropy serving as a critical indicator of training effectiveness.
Building on these observations, we propose \textbf{S}upervised \textbf{R}einforcement \textbf{F}ine-\textbf{T}uning (\method), a single-stage method that unifies both fine-tuning paradigms through entropy-aware weighting mechanisms. 
Our approach simultaneously applies SFT and RL to directly optimize the LLM using demonstrations and self-exploration rollouts rather than through two-stage sequential methods.
Extensive experiments show that \method achieves \textbf{59.1\%} average accuracy, outperforming zero-RL methods by \textbf{9.0\%} on five mathematical reasoning benchmarks and \textbf{10.9\%} on three out-of-distribution benchmarks.
\end{abstract}

\begin{quote}
	\vspace{-1em}
		\textbf{Project Website:} \url{https://anonymous.4open.science/w/SRFT2025} \vskip1ex
		\textbf{Model Website:} \url{https://huggingface.co/Yuqian-Fu/SRFT} 
	\vspace{-0.5em}
\end{quote}

\section{Introduction}

Recent advances in Large Language Models (LLMs) for reasoning~\citep{openai2025,guo2025deepseek,claude2025} have demonstrated remarkable capabilities in complex problem-solving tasks. 
Despite these remarkable achievements, fine-tuning strategies for enhancing reasoning capabilities remain an active area of research, presenting both opportunities and challenges.

Initial approaches often treat Supervised Fine-Tuning (SFT) and Reinforcement Learning (RL) as distinct, sequential phases. For instance, SFT might be used for instruction-following, followed by RL for alignment. However, this separation presents challenges: SFT can lead to models that memorize patterns without developing true reasoning abilities, potentially overfitting the training dataset~\citep{chu2025sft,chen2025sft}. Conversely, RL methods, while promising for exploration and reward optimization, can be sample inefficient, struggle with effective exploration in vast solution spaces~\citep{gao2025navigate,dou2025improving,schmied2025llms}, or suffer from issues like mode collapse, where the model repeatedly generates similar, suboptimal outputs~\citep{cai2025much}.

\begin{figure}[t]
	\vspace{-1.5em}
    \centering
	\subfigure[Toy illustration of SFT, RL, and \method for LLM reasoning training on a single prompt.]{
    \begin{minipage}[t]{0.465\textwidth}
        \centering
        \includegraphics[height=0.107\textheight]{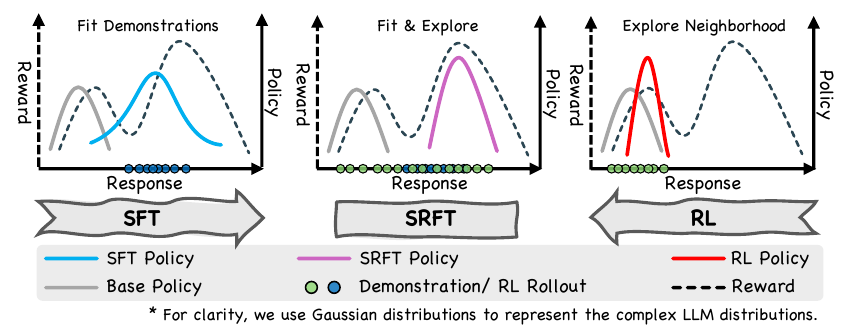}
        \label{fig:motivation}
    \end{minipage}
	}
	\subfigure[Framework of \method. Our method effectively leverages demonstrations to improve reasoning capabilities.]{
    \begin{minipage}[t]{0.5\textwidth}
        \centering
        \includegraphics[height=0.107\textheight]{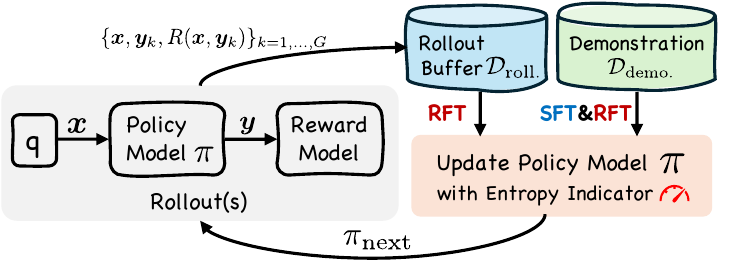}
        \label{fig:method}
    \end{minipage}
	}
    \caption{Overview of \method's motivation and framework.}
    \label{fig:motivation_method_results}
	\vspace{-1em}
\end{figure}

Instead of simple sequential approaches, recent work~\citep{yan2025learning,wu2025thought,liu2025uft,chen2025step,liu2025superrl} has shown a movement towards integrated frameworks that unify SFT and RL paradigms, or dynamically switch between the two fine-tuning methods during the LLM training process.
As illustrated in Figure~\ref{fig:motivation}, SFT guides LLM policies toward demonstration distributions, while RL enables policies to explore improved solutions in the neighborhood of the base policy. 
Our illustration demonstrates a special case: when the base policy is positioned near a suboptimal policy, the RL rollouts alone cannot effectively navigate to the optimal policy.
Beyond applying SFT and RL individually, the unified integration of SFT and RL within a single-stage method (e.g., our proposed \method) enables policies to directly optimize toward better solutions across an expanded space. 
However, a challenge remains in determining the balance between SFT's knowledge distillation and RL's policy optimization: insufficient integration risks error propagation and limits RL's improvements, while excessive reliance on demonstrations leads to overfitting that constrains exploration beyond the base policy distribution. 
This trade-off creates confusion for practitioners in choosing between SFT for leveraging demonstrations and RL for policy exploration.

To address these issues, in this work, we study how to build single-stage LLM fine-tuning algorithms that are not only effective for LLM reasoning from SFT datasets but also well-suited to continuous improvement with RL rollouts. We conduct a comprehensive analysis of the roles that SFT and RL play in fine-tuning LLM reasoning.
Through our analysis in Sec.~\ref{sec:analysis}, we obtain the following key findings that guide our subsequent algorithm design.
\newtcolorbox{promptbox}[1]{
  colback=blue!10,      
  colframe=black!75!black,    
  colbacktitle=black,         
  coltitle=white,             
  title=#1,                   
  boxrule=1pt,                
  arc=1mm,                    
  enhanced,
  attach boxed title to top left={xshift=3mm,yshift=-3mm},
  boxed title style={height=6mm},  
  left=3mm,
  right=3mm,
  top=3.25mm,
  bottom=1.75mm,
  width=0.98\textwidth,
  center,
}

\vspace{-0.3em}
\begin{promptbox}{Key Findings}
  \begin{itemize}[leftmargin=3.5mm,label=$\circ$]
    \item \textbf{Policy distribution effects} (Sec.~\ref{sec:token_distributions} and Sec.~\ref{sec:learning_dynamics}): During fine-tuning, SFT induces coarse-grained global changes to the LLM's policy distribution, while RL performs fine-grained selective modifications.
    \item \textbf{Single-stage optimization} (Sec.~\ref{sec:learning_dynamics} and Sec.~\ref{sec:single_stage_integration}): Single-stage integration of SFT and RL enables direct optimization for reasoning capabilities and achieves superior training efficiency compared to sequential SFT$\rightarrow$RL approaches.
    \item \textbf{Entropy as an indicator} (Sec.~\ref{sec:entropy_analysis}): Entropy dynamics reveal the underlying mechanisms of training processes, enabling balanced weighting between the two paradigms. 
  \end{itemize}
  \vspace{-0.5em}
\end{promptbox}

Based on these insights, we propose \textbf{S}upervised \textbf{R}einforcement \textbf{F}ine-\textbf{T}uning (\textbf{\method}), a single-stage method for LLM reasoning. As shown in Figure~\ref{fig:method}, we integrate SFT into RL and use entropy as an indicator to control the balance between these two paradigms. Specifically, for samples generated by LLM policy rollouts, we employ different RL training losses based on whether sample rewards are positive or negative. For samples from demonstration datasets, we simultaneously apply both SFT and RL objectives. This unified approach enables stable learning from demonstrations at multiple granularities while effectively bridging the complementary strengths of SFT and RL.

We evaluate our method on five competition-level mathematical reasoning benchmarks and three out-of-distribution (OOD) benchmarks. Our proposed \method achieves an accuracy of \textbf{59.1\%} based on Qwen-2.5-Math-7B~\citep{yang2024qwen2}, outperforming previous SFT and RL baselines by significant margins. Moreover, \method demonstrates superior generalization capability, achieving an average improvement of over \textbf{4.7\%} compared to other methods utilizing demonstrations.

Overall, \textbf{our key contributions} are:
\begin{itemize}[leftmargin=2.5em]
	\vspace{-0.5em}
    \item We conduct a comprehensive analysis of SFT and RL in LLM reasoning, examining their differential effects on policy distributions and learning dynamics. Besides, we analyze the integration of SFT and RL through an entropy-based lens.
    \item We propose \method, a single-stage fine-tuning approach that combines supervised fine-tuning and reinforcement learning with entropy-aware weighting mechanisms, enabling effective utilization of demonstrations while maintaining stable exploration dynamics.
    \item We demonstrate \method's superior performance across eight challenging benchmarks, achieving substantial improvements of \textbf{9.0\%} and \textbf{10.9\%} over zero-RL baselines on mathematical reasoning and out-of-distribution tasks, respectively.
	\vspace{-0.5em}
\end{itemize}

\section{Preliminaries}

\subsection{SFT and RL for LLM Reasoning}
\textbf{Supervised Fine-Tuning}
(SFT) is a standard approach for adapting pre-trained language models to specific downstream tasks or imparting particular stylistic characteristics. Given a dataset $\mathcal{D} = \{(\bm{x}_i, \bm{y}_i)\}_{i=1}^N$, where $\bm{x}_i$ is an input prompt and $\bm{y}_i$ is the corresponding target response generated by the behavior policy $\pi_\beta$, the objective is to train the language model policy $\pi_{\theta}$ (with parameters $\theta$) to maximize the conditional probability of generating the target response $\bm{y}_i$ given $\bm{x}_i$. This is typically achieved by minimizing the negative log-likelihood over the dataset:
\begin{equation}
	\label{eq:sft_loss}
	\mathcal{L}_\text{SFT}(\theta) = \mathbb{E}_{(\bm{x}, \bm{y}) \in \mathcal{D}} [-\log \pi_{\theta}(\bm{y} | \bm{x})],
\end{equation}
where $y_j$ is the $j$-th token in the response $\bm{y}$, and $\bm{y}_{<j}$ denotes the sequence of tokens in $\bm{y}$ before $y_j$.

\textbf{Reinforcement Learning}
(RL) is typically applied after SFT to further align LLMs with complex human preferences or desired behaviors (e.g., reasoning abilities, harmlessness) that are challenging to specify exhaustively through static datasets. In RL training, the LLM's token generation process is modeled as a Markov Decision Process (MDP)~\citep{puterman2014markov}.
We define a state $s_t$ at step $t$ as the concatenation of the input prompt $\bm{x}$ and all tokens generated so far $\bm{y}_{<t}$. This state serves as input to the policy model $\pi_{\theta}(\cdot|s_t)$. Specifically, the policy processes $s_t = (\bm{x}, \bm{y}_{<t}) = (x_1, x_2, \ldots, x_l, y_1, y_2, \ldots, y_{t-1})$, where $x_i$ denotes the $i$-th token of the input $\bm{x}$ and $y_j$ represents the token generated by $\pi_{\theta}$ at step $j$. An action $a_t$ corresponds to the selection of the next output token $y_t$. 
The LLM, acting as a policy $\pi_{\theta}(a_t|s_t)$, generates a trajectory $\bm{y}$ (a sequence of tokens) in response to the prompt $\bm{x}$. A reward function $R(\bm{x}, \bm{y})=\sum_{t=1}^{T} r(\bm{x},y_t)$ provides a scalar score for the entire trajectory $\bm{y}$ given prompt $\bm{x}$, typically derived from human evaluations or automated metrics. 
In the context of RL, the behavior policy $\pi_{\beta}(\bm{y}|\bm{x})$ refers to the model that generated the responses in the replay buffer. This policy is crucial for RL, particularly for off-policy learning, as it enables proper importance sampling corrections to account for the distribution shift between the data-generating model and the current training model.
The MDP formulation in LLMs presents several notable characteristics:
\begin{itemize}[leftmargin=2.5em]
    \item \textbf{Sequential state representation:} At each step $t$, the state $s_t \in S$ consists of the concatenation of the input prompt $\bm{x}$ and all actions (tokens) generated so far $\bm{y}_{<t}$. This state serves as input to the policy model $\pi_{\theta}(\cdot|s_t)$.
    \item \textbf{Sparse and delayed rewards:} Rewards $R(\bm{x}, \bm{y})$ are typically sparse, provided only upon completion of a sequence $\bm{y}$. This dependency on the final output's overall quality complicates credit assignment across the generation process.
\end{itemize}

\subsection{Policy Optimization in Reinforcement Learning}
\label{sec:rft_pre}
To optimize the LLM policy, Group Relative Policy Optimization (GRPO)~\citep{shao2024deepseekmath} offers a different RL algorithm that is presented as a memory-efficient variant of Proximal Policy Optimization (PPO)~\citep{schulman2017proximal}. A key characteristic is that GRPO typically operates without a learned value function. Instead, for a given prompt $\bm{x}$, it often generates a group of $G$ responses $\{\bm{y}_1, \ldots, \bm{y}_G\}$ using the current policy. The rewards $\{R(\bm{x}, \bm{y}_1), \ldots, R(\bm{x}, \bm{y}_G)\}$ for these responses are then used to compute a relative advantage for each response:
\begin{equation}
	\label{eq:advantage}
	\hat{A}_{k} = \frac{R(\bm{x}, \bm{y}_k) - \text{mean}(\{R(\bm{x}, \bm{y}_k)|k=1,2,\ldots,G\})}{\text{std}(\{R(\bm{x}, \bm{y}_k)|k=1,2,\ldots,G\})}.
\end{equation}

Then, GRPO maximizes a clipped surrogate objective function to ensure stable updates. Let $\pi_{\theta_{\text{old}}}$ be the policy before the update. For each token $y_{k,t}$ in a trajectory $\bm{y}_k$ (from state $s_{t}$), the importance sampling ratio is $r_{k,t}(\theta) = \frac{\pi_{\theta}(y_{k,t}|s_{t})}{\pi_{\theta_{\text{old}}}(y_{k,t}|s_{t})}$. The objective function for GRPO can then be expressed as:
\begin{equation}
	\mathcal{J}_\text{GRPO}(\theta) = \frac{1}{G} \sum_{k=1}^{G} \frac{1}{|\bm{y}_k|} \sum_{t=1}^{|\bm{y}_k|} \left[\min\left\{r_{k,t}(\theta)\cdot\hat{A}_{k}, \texttt{clip}\left\{r_{k,t}(\theta), 1-\epsilon, 1+\epsilon\right\}\cdot\hat{A}_{k}\right\}\right], 
\end{equation}
where $\epsilon$ is a small hyperparameter that defines the clipping range. Through this mechanism, the LLM policy is updated while maintaining stable gradient constraints.

\section{Analysis of SFT and RL in LLM Reasoning}
\label{sec:analysis}

In this section, we provide a comprehensive analysis of the roles of Supervised Fine-Tuning (SFT) and Reinforcement Learning (RL) in LLM reasoning. We first examine their differential effects on token distributions (Sec.~\ref{sec:token_distributions}) and learning dynamics (Sec.~\ref{sec:learning_dynamics}), then investigate their integration mechanisms through an entropy-based perspective (Sec.~\ref{sec:sft_rft_integration}). All experiments are conducted across five mathematical reasoning benchmarks (AIME24, AMC, MATH500, Minerva, and Olympiad) with results averaged. 
We tune hyperparameters for all baseline methods to ensure fair and optimal performance comparisons.

\subsection{SFT and RL Effects on LLMs: Sledgehammer vs. Scalpel}

\subsubsection{Effects on Token Distributions}
\label{sec:token_distributions}

\begin{figure}[t]
	\vspace{-2em}
	\centering
	\includegraphics[width=\textwidth]{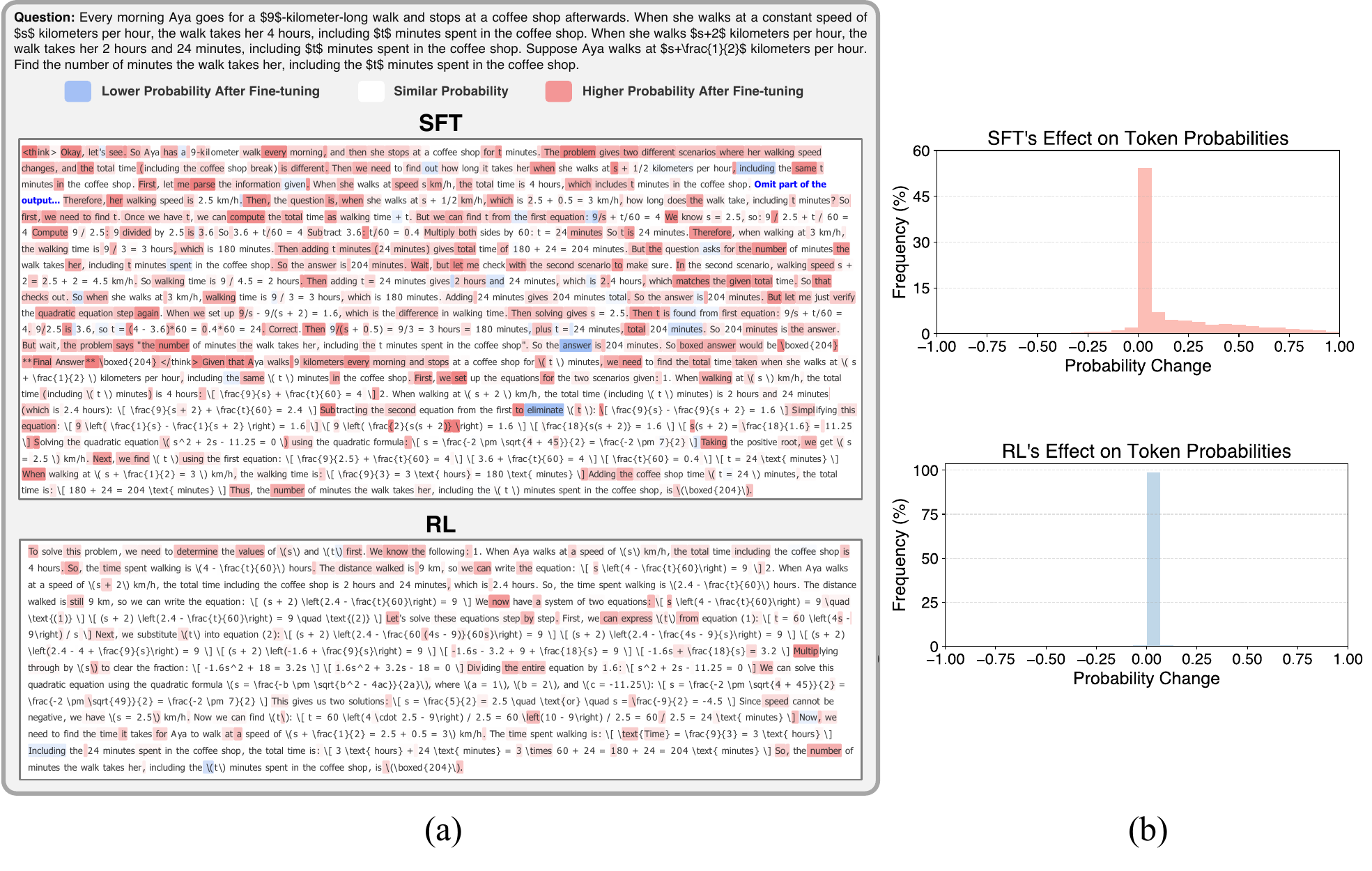}
	\vspace{-2em}
	\caption{Visualization of LLM distribution changes during fine-tuning. (a) Heatmap visualization comparing responses generated by fine-tuned and base models, where darker background colors indicate larger probability changes. (b) Distribution of token probability changes across five mathematical reasoning benchmarks.}
	\label{fig:token_prob}
	\vspace{-0.5em}
\end{figure}
To understand the differential impact of SFT and RL for reasoning, we visualize probability changes across response tokens to identical prompts before and after fine-tuning the same base model (Qwen-2.5-Math-7B).
As illustrated in Figure~\ref{fig:token_prob}(a), the results reveal a fundamental asymmetry: SFT substantially alters the probability distribution across the entire response sequence, while RL selectively modifies probabilities for only a small subset of tokens, while leaving numerical content and mathematical proof statements largely unchanged.
We further quantify these distribution shifts across five benchmarks, as shown in Figure~\ref{fig:token_prob}(b). The results demonstrate that SFT produces more pronounced changes to policy distributions compared to RL, with token probability changes in RL clustering near zero while SFT exhibits substantially larger magnitude shifts. From a theoretical perspective, this behavior can be understood through the gradient of the SFT objective function:
\begin{equation}
\nabla_\theta \mathcal{L}_{\text{SFT}} = \mathbb{E}_{(\bm{x}, \bm{y}) \sim \mathcal{D}} \left[ \sum_{t=1}^{|\bm{y}|} \sum_{v \in \mathcal{V}} \left( \pi_\theta(v|\bm{x}, \bm{y}_{<t}) - \mathbf{1}_{v = y_t} \right) \nabla_\theta \log \pi_\theta(v|\bm{x}, \bm{y}_{<t}) \right],
\end{equation}
where $\mathcal{V}$ is the LLM vocabulary, and $\mathbf{1}_{v = y_t}$ is an indicator function that equals 1 when token $v$ matches the target token $y_t$ and 0 otherwise. The detailed derivation is provided in Appendix~\ref{sec:sft_gradient_derivation}. This formulation reveals that SFT systematically sharpens the model distribution by increasing probabilities for target tokens while decreasing probabilities for all other tokens in the vocabulary, leading to more deterministic outputs.

\subsubsection{Visualization of Learning Dynamics}
\label{sec:learning_dynamics}

\begin{wrapfigure}[21]{r}{0.45\textwidth}
	\centering
	\includegraphics[width=0.45\textwidth]{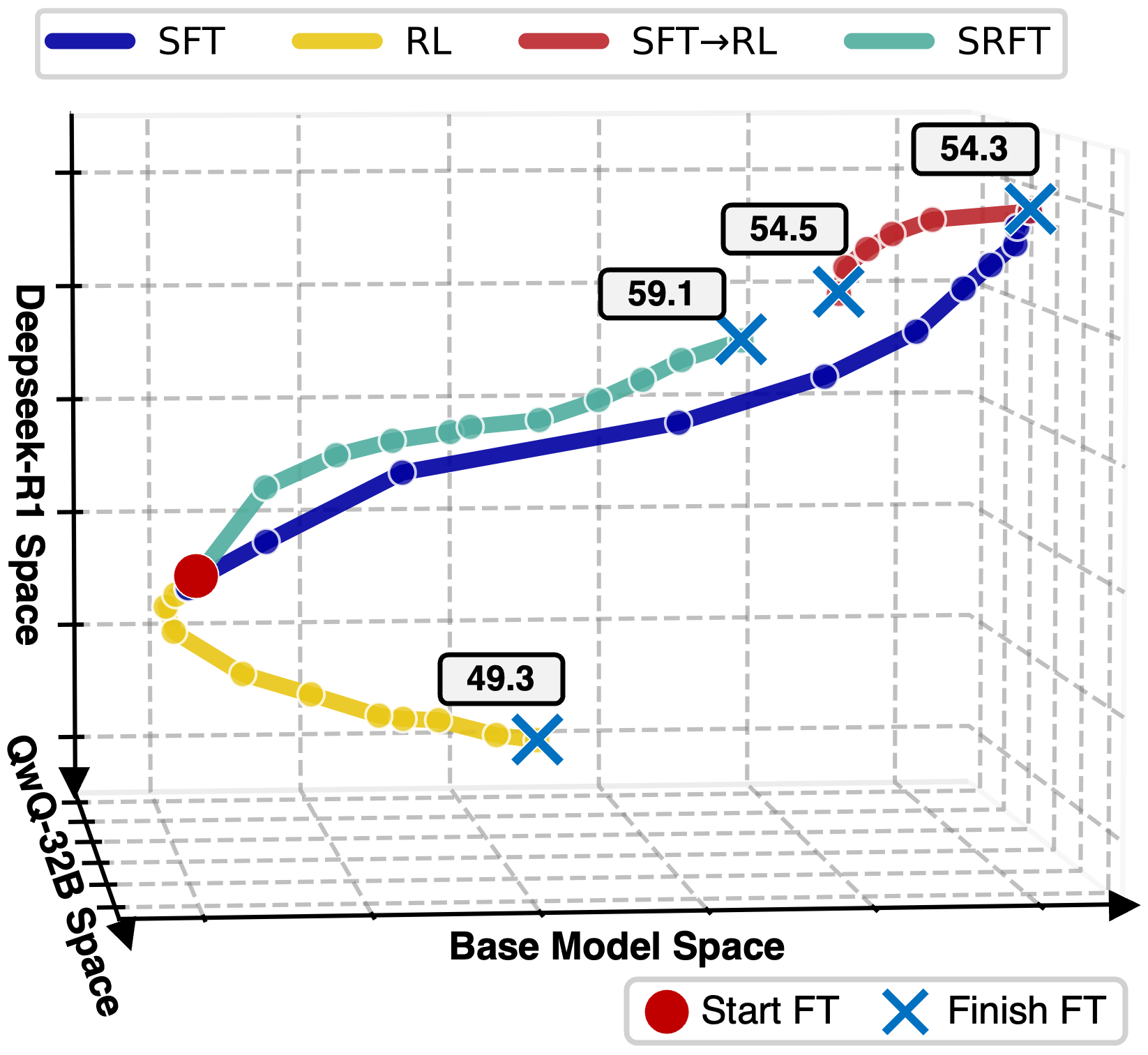}
	\vspace{-1.5em}
	\caption{Learning dynamics during different fine-tuning paradigms in three-dimensional probability space.
	The \tcbox[box align=base,nobeforeafter,size=small,left=0pt,right=0pt,top=0pt, bottom=0pt,boxsep=1pt,arc=0.5mm,colback=black!0,colframe=black!100,boxrule=1pt]{number} denotes the final performance of each training process.
	}
	\label{fig:trajectory}
\end{wrapfigure}
Beyond the token probability analysis, we analyze the training paradigms from the perspective of learning dynamics. Since directly measuring the LLM feature space is computationally intractable, we propose a novel visualization approach that \textit{maps each model to a point in the vocabulary probability space}, treating models as functions that transform prompts to output probability distributions over the vocabulary. We establish three reference models—the base model (Qwen-2.5-Math-7B), DeepSeek-R1, and QwQ-32B~\citep{qwq32b}—as coordinate frames, enabling indirect measurement of model evolution at different fine-tuning steps through the distance in probability space (two models are considered \textit{close} if they assign similar output probabilities to all tokens across all prompts). 
Detailed methodology for this visualization is provided in Appendix~\ref{sec:model_trajectory}.

The visualization is shown in Figure~\ref{fig:trajectory}, which demonstrates that all fine-tuning paradigms exhibit performance improvements while simultaneously moving away from the base model space (Qwen-2.5-Math-7B). 
Specifically, SFT exhibits greater distribution changes from the base model compared to RL and achieves higher performance, which further validates our observations in Sec.~\ref{sec:token_distributions} that SFT induces larger changes to model distributions while RL fine-tuning within a neighborhood of the initialization point.
We further analyze two integration approaches: the two-stage SFT$\rightarrow$RL method and our proposed single-stage \method approach detailed in the following section. 
The results reveal that the learning dynamics of the two-stage SFT$\rightarrow$RL method traverse from the post-SFT model toward a higher-performance region that paradoxically lies closer to the base model, suggesting that the initial SFT phase may induce excessive distributional deviation from the base model, thereby compromising the effectiveness of subsequent RL.
In contrast, our single-stage method demonstrates more constrained yet targeted changes in the probability space, enabling more precise optimization compared to sequential integration approaches.

\subsection{Integration of SFT and RL: From Two-stage to Single-stage}
\label{sec:sft_rft_integration}

\subsubsection{Sequential Integration Analysis}
\label{sec:entropy_analysis}

\begin{table}[t]
    \centering
    \renewcommand{\arraystretch}{1.1}
    \caption{Performance comparison of SFT and RL integration strategies across multiple benchmarks. 
    \textbf{Bold} and \underline{underlined} indicate the best and second-best performance, respectively.}
    \label{tab:sft_rft_integration}
    \resizebox{0.85\textwidth}{!}{%
    \begin{tabular}{@{}lrrrrrr@{}}
    \toprule
    \textbf{Model} &
      \multicolumn{1}{c}{\textbf{AIME24}} &
      \multicolumn{1}{c}{\textbf{AMC}} &
      \multicolumn{1}{c}{\textbf{MATH500}} &
      \multicolumn{1}{c}{\textbf{Minerva}} &
      \multicolumn{1}{c}{\textbf{Olympiad}} &
      \multicolumn{1}{c}{\textbf{Avg.}} \\ \midrule
    Qwen2.5-Math-7B & 14.1 & 44.8 & 64.8 & 16.5 & 29.6 & 34.0 \\
    SFT             & \underline{21.2} & \underline{53.2} & 83.0 & \underline{37.1} & 42.2 & 47.3 \\
    RL             & \underline{21.2} & \textbf{59.3} & \underline{83.6} & 36.4 & \underline{46.6} & \underline{49.4} \\
    RL$\rightarrow$SFT         & 10.5 & 40.4 & 73.6 & 32.0 & 30.7 & 37.4 \\
    RL$\rightarrow$SFT$_{\text{KL}}$     & 13.1 & 45.2 & 70.2 & 26.5 & 36.3 & 38.3 \\
    SFT$\rightarrow$RL         & \textbf{24.5} & \textbf{59.3} & \textbf{86.4} & \textbf{39.3} & \textbf{53.1} & \textbf{52.5} \\ \bottomrule
    \end{tabular}%
    }
    \vspace{-0.5em}
    \end{table}

\begin{figure}[t]
    \centering
	\subfigure[Reasoning capability comparison between SFT$\rightarrow$RL and RL$\rightarrow$SFT.]{
        \includegraphics[height=0.2\textheight]{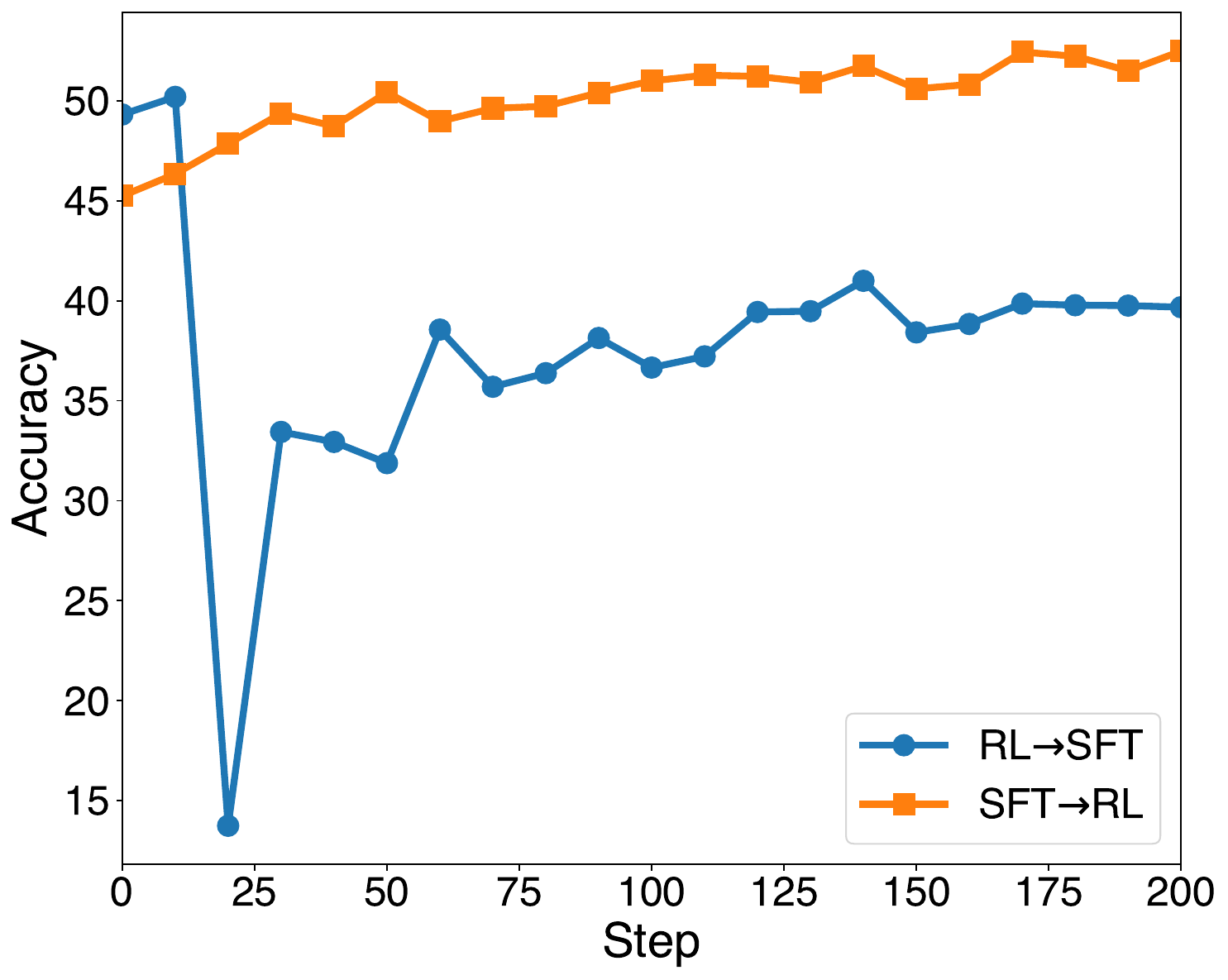}
        \label{fig:performance_comparison}
    }
    \hspace{0.03\textwidth}
	\subfigure[Entropy dynamics of two different fine-tuning paradigms.]{
        \includegraphics[height=0.2\textheight]{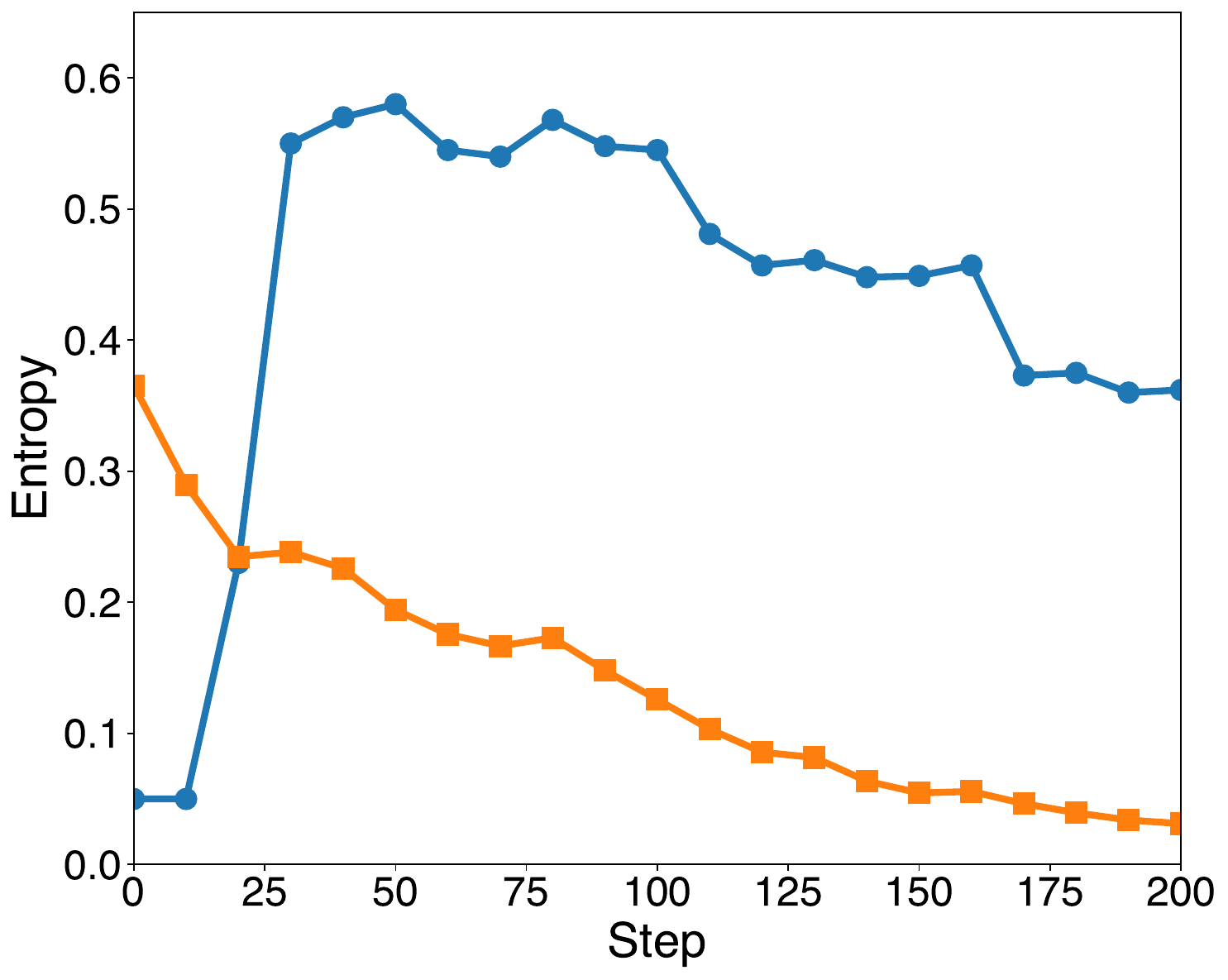}
        \label{fig:entropy_dynamics}
    }
	\vspace{-0.5em}
	\caption{Comparison between SFT$\rightarrow$RL and RL$\rightarrow$SFT.}
	\label{fig:SFT-RFT}
	\vspace{-1.5em}
\end{figure}

In this section, we examine the integration of SFT and RL through the lens of entropy dynamics to understand their complementary roles in LLM fine-tuning. We begin by systematically analyzing two sequential integration approaches: SFT$\rightarrow$RL and RL$\rightarrow$SFT, as shown in Figure~\ref{fig:SFT-RFT}. 

As demonstrated in Table~\ref{tab:sft_rft_integration} and Figure~\ref{fig:performance_comparison}, applying SFT after RL consistently yields suboptimal performance across all benchmarks. 
To mitigate the detrimental policy shifts induced by RL$\rightarrow$SFT, we introduced a KL divergence constraint ($\text{SFT}_{\text{KL}}$, detailed in Appendix~\ref{sec:experimental_details}) to regularize the distribution changes. However, even with this constraint, the performance improvements remained limited, suggesting fundamental incompatibility in this ordering.
In contrast, existing methods successfully achieve substantial performance gains through RL when applied after the base model SFT, as evidenced in Table~\ref{tab:sft_rft_integration}. This asymmetric behavior reveals that the sequence of fine-tuning paradigms critically affects the final model performance, motivating our entropy-based analysis to understand the underlying mechanisms.

To understand this asymmetric behavior, we analyze the training dynamics of SFT and RL from \textit{an entropy perspective}. As illustrated in Figure~\ref{fig:entropy_dynamics}, policies after RL exhibit significantly lower entropy, approaching deterministic outputs. However, the distribution shift introduced by subsequent SFT causes a rapid increase in entropy (corresponding to the sharp performance drop in Figure~\ref{fig:performance_comparison}), followed by a gradual decline.
Moreover, models after RL demonstrate a limited capacity for further learning through SFT, as evidenced by the entropy plateau occurring after approximately 90 training steps (Figure~\ref{fig:entropy_dynamics}). In contrast, base models undergoing SFT exhibit a brief initial entropy increase followed by a sustained decrease, ultimately yielding performance improvements. This distinct entropy trajectory suggests that while RL effectively enhances LLM performance, it simultaneously reduces the model's plasticity—its capacity to adapt through subsequent training. These findings establish entropy as a crucial indicator for effective SFT and RL integration.

\subsubsection{Single-Stage Integration Analysis}
\label{sec:single_stage_integration}

\begin{wrapfigure}[17]{r}{0.46\textwidth}
	\vspace{-1em}
	\centering
	\includegraphics[width=0.46\textwidth]{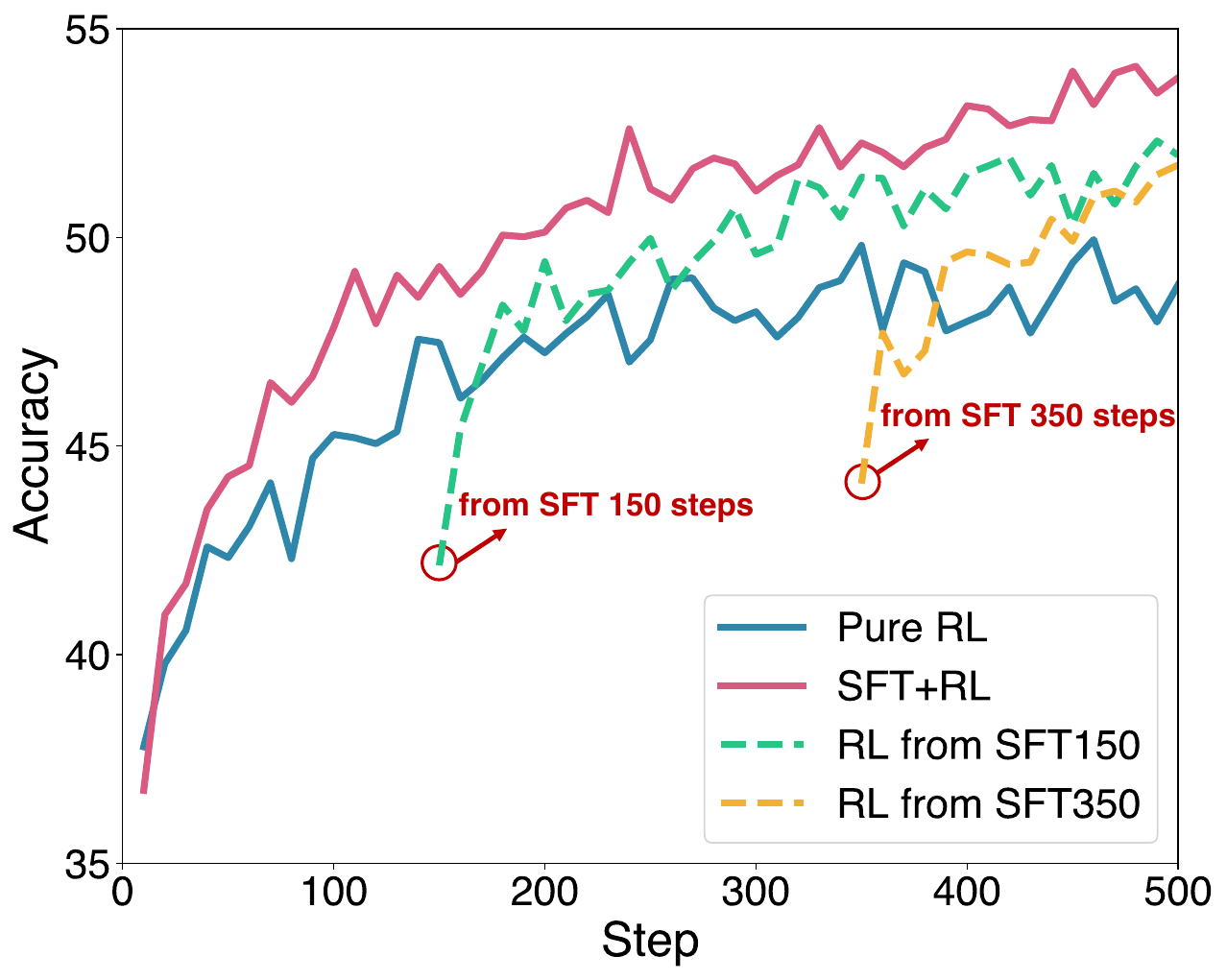}
	\vspace{-2.5em}
	\caption{Preliminary comparison across pure RL, sequential SFT$\rightarrow$RL, and single-stage SFT+RL integration approaches.}
	\label{fig:SFT-RFT-performance}
\end{wrapfigure}
Building upon the analysis above, we establish that the SFT$\rightarrow$RL paradigm demonstrates superior suitability for LLM reasoning compared to RL$\rightarrow$SFT. However, beyond these sequential integration approaches, we investigate a single-stage approach that directly unifies both paradigms (SFT+RL), with the combined objective $\mathcal{L}_{\text{SFT+RL}} = \mathcal{L}_{\text{SFT}} + \mathcal{L}_{\text{RL}}$.
We conduct a preliminary experiment comparing pure RL, sequential SFT$\rightarrow$RL with varying SFT steps, and single-stage SFT+RL, as illustrated in Figure~\ref{fig:SFT-RFT-performance}.
Our empirical findings reveal that the single-stage SFT+RL method achieves superior training efficiency compared to the sequential SFT$\rightarrow$RL approach.
Notably, we observe an intriguing phenomenon in models with extensive SFT pre-training (350 steps SFT followed by 150 steps RL): a transient performance degradation during the initial phases of RL. 
We attribute this behavior to two principal factors:
First, SFT datasets derived from other models' responses may not consistently represent optimal solutions, even when sourced from high-quality demonstrations, potentially leading to suboptimal policy learning during the SFT phase. Second, pure RL exhibits limited data efficiency due to its inability to effectively leverage demonstrations. In the sequential SFT$\rightarrow$RL training paradigm, the RL phase may induce catastrophic forgetting of knowledge acquired during SFT~\citep{cai2025much}, resulting in transient performance deterioration.
In contrast, the single-stage SFT+RL method effectively leverages demonstrations through unified optimization. This approach enables direct policy optimization toward the target objective while preserving the knowledge distillation benefits of supervised learning from datasets.
Importantly, both datasets utilization methods significantly outperform pure RL across all performance metrics.

\section{Method}

In this section, we present the Supervised and Reinforcement Fine-tuning (\method) algorithm, which integrates the advantages of Supervised Fine-Tuning (SFT) and Reinforcement Learning (RL) in a single-stage approach. 
Building upon the RL framework described in Sec.~\ref{sec:rft_pre}, \method incorporates flexible guidance from demonstrations, enabling the algorithm to harness the complementary strengths of both fine-tuning paradigms. The core innovation of \method lies in its single-stage learning mechanism: coarse-grained behavior policy approximation through SFT and fine-grained policy refinement through RL, both applied to demonstration data and self-generated trial-and-error data.

\subsection{Learning from Demonstrations}

Given a dataset containing demonstrations $\mathcal{D}_\text{demo.}$ (e.g., reasoning responses generated by DeepSeek-R1), \method employs a dual-pronged strategy to effectively harness this valuable resource. First, we leverage SFT to perform a coarse-grained approximation of the behavior policy underlying the expert's responses. The behavior policy $\pi_{\beta}(\bm{y}|\bm{x})$ captures the underlying generation patterns that produced these high-quality responses, which we seek to approximate through supervised learning:
\begin{equation}
 \mathcal{L}^{\text{demo.}}_{\text{SFT}}=\mathbb{E}_{(\bm{x}, \bm{y}) \sim \mathcal{D}_\text{demo.}}[-\log \pi_{\theta}(\bm{y}|\bm{x})].
\end{equation}
Second, we adopt an off-policy RL approach similar to LUFFY~\citep{yan2025learning} to perform fine-grained learning of the behavior policy through RL. Specifically, we directly augment the LLM's on-policy rollout group with demonstrations, creating a heterogeneous training batch:
\begin{equation}
G_{\text{aug.}} = \{(\bm{x}_i, \bm{y}_i)\}_{i=1}^{|G_{\text{roll.}}|} \cup \textcolor[HTML]{c04851}{\{(\bm{x}_j, \bm{y}_j)\}_{j=1}^{|G_{\text{demo.}}|}},
\end{equation}
where $G_{\text{roll.}}$ denotes the on-policy rollout group, $G_{\text{demo.}}$ denotes the demonstration group.
The advantage estimation for the entire group is given by:
\begin{equation}
\hat{A}_k  = \frac{r(\bm{x}, \bm{y}_k) - \text{mean}(\{r(\bm{x}, \bm{y}_k)|k=1,2,\ldots,\textcolor[HTML]{c04851}{|G_{\text{aug.}}|}\})}{\text{std}(\{r(\bm{x}, \bm{y}_k)|k=1,2,\ldots,\textcolor[HTML]{c04851}{|G_{\text{aug.}}|}\})}.
\end{equation}
Since responses generated by expert LLMs typically exhibit higher rewards, their inclusion increases the advantage estimation for the entire group as shown in Eq.~\eqref{eq:advantage}, promoting optimistic exploration in the LLM policy through this mechanism.

To address the distribution mismatch between behavior policies $\pi_{\beta}$ of demonstrations and the current training policy $\pi_{\theta}$ identified in our analysis, we implement two key mitigation strategies:
\begin{itemize}[leftmargin=2.5em]
	\item For SFT on demonstrations, our entropy analysis in Sec.~\ref{sec:sft_rft_integration} demonstrates that entropy serves as a crucial indicator for effective SFT and RL integration. Motivated by this insight, we introduce an adaptive weighting mechanism that dynamically adjusts based on the current policy entropy, employing $w_\text{SFT} = 0.5* \texttt{stop\_grad}(\exp(-\mathcal{H}(\pi_{\theta})))$ as the SFT weight, where $\texttt{stop\_grad}(\cdot)$ denotes the gradient stopping operation. This entropy-aware mechanism ensures that when the policy exhibits high entropy (indicating uncertainty), the SFT training loss exerts diminished influence on the model updates, thereby mitigating performance degradation caused by distribution mismatch between the behavior policy of demonstrations and the current policy while still enabling effective behavior policy approximation:
\begin{equation}
\label{eq:sft_external}
\mathcal{L}_{\text{SFT}}^{\text{demo.}}(\theta) = w_\text{SFT} \cdot \mathbb{E}_{(\bm{x}, \bm{y}) \sim \mathcal{D}_{\text{demo.}}}[-\log \pi_{\theta}(\bm{y}|\bm{x})].
\end{equation}
	\item For off-policy RL training, we introduce an importance sampling term similar to GRPO~\citep{shao2024deepseekmath} and PPO~\citep{schulman2017proximal} to account for the distribution shift between the behavior policy and the current policy:
\begin{equation}
\mathcal{L}_{\text{RL}}^{\text{demo.}}(\theta) = -\mathbb{E}_{\textcolor[HTML]{c04851}{(\bm{x}, \bm{y}) \sim \mathcal{D}_{\text{demo.}}}} \left[ \min\left\{r_{k,t}(\theta) \cdot \hat{A}_k, \texttt{clip}\left\{r_{k,t}(\theta), 1-\epsilon, 1+\epsilon\right\} \cdot \hat{A}_k\right\} \right],
\end{equation}
\begin{equation}
r_{k,t}(\theta) = \frac{\pi_{\theta}(\bm{y}_{k,t}|\bm{x}_t)}{\pi_{\beta}(\bm{y}_{k,t}|\bm{x}_t)}.
\end{equation}
\end{itemize}
Following practices established in recent work~\citep{yan2025learning,ma2025learning}, we set the behavior policy $\pi_{\beta} = 1$ to avoid tokenization complexities that arise when aligning the current training policy with the behavior policy, thereby facilitating easy integration of off-the-shelf datasets without requiring recomputation of behavior policy probabilities. Additionally, we omit the clipping operation, as the standard clipping mechanism becomes imbalanced and potentially unstable when $\pi_{\beta} = 1$.

\subsection{Learning from Self-Exploration}

In addition to leveraging the demonstration data, \method enables the LLM policy to learn simultaneously from its own exploration rollouts. While traditional RL methods learn from both positive and negative samples generated during rollouts, we observe that under on-policy RL with binary rewards $\{1, -1\}$, the basic RL objective function can be naturally decomposed into two distinct components:
\begin{equation}
\begin{aligned}
\mathcal{L}_{\text{RL}}^{\text{self-rollout}} &= -\mathbb{E}_{\bm{x} \sim \mathcal{D}, \bm{y} \sim \pi_{\theta}(\cdot|\bm{x})} [R(\bm{x}, \bm{y}) \log \pi_{\theta}(\bm{y}|\bm{x})] \\
&= \underbrace{\mathbb{E}_{\bm{x}\sim \mathcal{D},\bm{y}^+ \sim \pi_{\theta}(\cdot|\bm{x})} [-\log \pi_{\theta}(\bm{y}^+|\bm{x})]}_{\text{Positive Sample }\Circled{1}} + \underbrace{\mathbb{E}_{\bm{x}\sim \mathcal{D},\bm{y}^- \sim \pi_{\theta}(\cdot|\bm{x})} [\log \pi_{\theta}(\bm{y}^-|\bm{x})]}_{\text{Negative Sample }\Circled{2}},
\end{aligned}
\end{equation}
where $\mathcal{D}$ denotes the RL training dataset, and $\bm{y}^+$ and $\bm{y}^-$ represent the correct and incorrect responses, respectively.
A critical insight emerges from this decomposition: the positive sample objective $\Circled{1}$ exhibits structural similarity to supervised fine-tuning, as it maximizes the likelihood of correct responses. However, these positive samples are generated on-policy by the current policy $\pi_{\theta}$ rather than sourced from SFT datasets, distinguishing our approach from conventional supervised learning paradigms. The negative reward component $\Circled{2}$ implements likelihood minimization, systematically reducing the probability mass assigned to incorrect responses.  This structural correspondence suggests that learning from positive samples constitutes a coarse-grained optimization strategy that necessitates careful balance.
Moreover, in contrast to learning from demonstrations, self-exploration induces rapid entropy reduction as the model converges toward increasingly deterministic outputs, potentially compromising exploration capabilities. To mitigate this phenomenon and preserve training stability, inspired by our analysis in Sec.~\ref{sec:token_distributions}, we introduce an entropy-adaptive weighting mechanism $w_\text{RL} = 0.1*\texttt{stop\_grad}(\exp(\mathcal{H}(\pi_{\theta})))$ specifically for the positive sample objective. This mechanism is similar to our formulation in Eq.~\eqref{eq:sft_external} but serves the complementary purpose of maintaining exploration diversity. The complete self-exploration objective is formulated as:
\begin{equation}
\mathcal{L}_{\text{RL}}^{\text{self-rollout}}(\theta) = \textcolor[HTML]{c04851}{w_\text{RL}} \cdot \mathbb{E}_{\bm{x}\sim \mathcal{D},\textcolor[HTML]{c04851}{\bm{y}^+ \sim \pi_{\theta}(\cdot|\bm{x})}} [-\log \pi_{\theta}(\bm{y}^+|\bm{x})] + \mathbb{E}_{\bm{x}\sim \mathcal{D},\textcolor[HTML]{c04851}{\bm{y}^- \sim \pi_{\theta}(\cdot|\bm{x})}} [\log \pi_{\theta}(\bm{y}^-|\bm{x})].
\end{equation}

\subsection{Integrating Demonstrations with Self-Exploration Rollouts in a Single-Stage Approach}

By leveraging both demonstrations and self-generated rollouts, \method effectively balances the coarse-grained adjustments of SFT with the fine-grained refinements of RL throughout the single-stage fine-tuning process. The total loss function combines all four components:
\begin{equation}
\mathcal{L}_{\text{\method}}(\theta) = \mathcal{L}_{\text{SFT}}^{\text{demo.}}(\theta) + \mathcal{L}_{\text{RL}}^{\text{demo.}}(\theta) + \mathcal{L}_{\text{RL}}^{\text{self-rollout}}(\theta).
\end{equation}
This objective enables \method to simultaneously benefit from demonstrations and self-exploration rollouts while maintaining stable training dynamics through two entropy-aware weighting mechanisms.

\section{Experiments.}

\subsection{Experimental Setups}

\paragraph{Training Datasets.}
We employ OpenR1-Math-46k-8192\footnote{\url{https://huggingface.co/datasets/Elliott/Openr1-Math-46k-8192}}~\citep{yan2025learning} as the training dataset for \method, which constitutes a subset of OpenR1-Math-220k~\citep{openr1} comprising 46,000 mathematical problems sourced from NuminaMath 1.5~\citep{numina_math_datasets}, accompanied by high-quality reasoning responses generated by DeepSeek-R1~\citep{guo2025deepseek}. The dataset undergoes filtering through Math-Verify\footnote{\url{https://github.com/huggingface/Math-Verify}} to exclude instances with unverifiable answers or responses exceeding 8,192 tokens in length. This dataset serves multiple purposes in our framework: providing prompts for policy rollouts, ground-truth answers for reward computation, and high-quality demonstrations for \method.
The dataset details are provided in Appendix~\ref{sec:dataset_details}.

\paragraph{Evaluation.}
We conduct a comprehensive evaluation on several widely-adopted mathematical reasoning benchmarks, including AIME24~\citep{li2024numinamath}, AMC~\citep{li2024numinamath}, Minerva~\citep{lewkowycz2022solving}, OlympiadBench~\citep{he2024olympiadbench}, and MATH500~\citep{hendrycks2021measuring}. For datasets with limited sample sizes (AIME24 and AMC), we report the avg@32 metric; for the remaining three datasets, we adopt pass@1 as the evaluation criterion. 
Given that our method primarily focuses on mathematical reasoning capabilities, we further assess the model's generalization ability on three out-of-distribution benchmarks: ARC-C~\citep{clark2018think} (open-domain reasoning), 
GPQA-Diamond~\citep{rein2024gpqa} (graduate-level scientific knowledge, denoted as GPQA-D), and MMLU-Pro~\citep{wang2024mmlu} (reasoning problems from academic examinations and textbooks).
To mitigate potential information leakage, we randomly shuffle the option orders for all multiple-choice questions.
During inference, we set the generation temperature to 0.6 with a maximum response length of 8,192 tokens. 
We employ Math-verify as the verifier for training validation and the OAT-Grader~\citep{liu2024oat} for final evaluation.

\paragraph{Baseline Methods.}
We benchmark \method against the following baselines using Qwen2.5-Math-7B as the base model. \textbf{SFT methods:} (1) SFT on OpenR1-Math-46k-8192 dataset; (2) SFT training with KL divergence constraints incorporated into the loss function (SFT$_{\text{KL}}$). \textbf{RL methods:} (3) RL$_\text{GRPO}$~\citep{shao2024deepseekmath}, a simplified PPO variant trained on the same 46k dataset; 
(4) Simple-RL-Zero~\citep{zeng2025simplerl}, applying GRPO to approximately 24k mathematical samples from GSM8K and MATH; (5) OpenReasoner-Zero~\citep{hu2025open}, a PPO-based approach trained on 129k multi-source samples including AIME; (6) PRIME-Zero~\citep{cui2025process}, conducting policy rollouts on 150k NuminaMath queries with implicit process rewards and final labels. 
\textbf{SFT and RL methods:} (7) SFT\textrightarrow{}RL, sequential training with SFT the same 46k dataset followed by GRPO; 
(8) ReLIFT~\citep{ma2025learning}, an approach that interleaves RL with online Fine-Tuning on the hardest questions;
(9) LUFFY~\citep{yan2025learning}, a mixed-policy GRPO approach using the same 46k dataset; (10) TAPO~\citep{wu2025thought}, dynamically integrating structured external knowledge within the GRPO framework, trained on 5.5k samples from the MATH dataset.

\paragraph{Implementation Details.}
Following recent work~\citep{yan2025learning,wu2025thought,cui2025process}, we use the Qwen2.5-Math-7B~\citep{yang2024qwen2} model as the base model. 
In \method, we generate 8 rollout trajectories per prompt with a maximum sequence length of 8,192 tokens. All experiments are conducted over 500 training steps. Comprehensive experimental details are provided in Appendix~\ref{sec:experimental_details}.

{\addtolength{\extrarowheight}{\belowrulesep}
\addtolength{\extrarowheight}{\aboverulesep}

\aboverulesep=0pt
\belowrulesep=0pt
\begin{table}[t]
    \caption{Overall performance on five competition-level mathematical reasoning benchmarks and three out-of-distribution benchmarks based on Qwen2.5-Math-7B. \textbf{Bold} and \underline{underlined} indicate the best and second-best performance, respectively.}
    \label{tab:main_performance}
    \renewcommand{\arraystretch}{1.1}
    \centering
    \resizebox{\textwidth}{!}{%
    \begin{tabular}{>{}lcccccc|cccc<{}}
    \toprule
     &
     \multicolumn{6}{c|}{\cellcolor[HTML]{FFF3CC}{\textbf{In-Distribution  Performance}}} &
      \multicolumn{4}{c}{\cellcolor[HTML]{FFF3CC}{\textbf{Out-of-Distribution  Performance}}} \\ \cmidrule(l){2-11} 
    \multirow{-2}{*}{\textbf{Model}} &
      \textbf{AIME24} &
      \textbf{AMC} &
      \textbf{MATH500} &
      \textbf{Minerva} &
      \textbf{Olympiad} &
      \textbf{Avg.} &
      \textbf{ARC-C} &
      \textbf{GPQA-D} &
      \textbf{MMLU-Pro} &
      \textbf{Avg.} \\ \midrule
    \textbf{Qwen2.5-Math} &
      11.4 &
      32.6 &
      48.8 &
      8.7 &
      15.8 &
      23.5 &
      18.2 &
      11.1 &
      16.9 &
      15.4 \\
    \textbf{Qwen2.5-Math-Instruct} &
      12.9 &
      48.3 &
      81.2 &
      33.1 &
      39.8 &
      43.1 &
      70.3 &
      24.7 &
      34.1 &
      43.0 \\ \midrule
      {\color[HTML]{656565}\textbf{\textit{Supervised Fine-Tuning}}} &
       &
       &
       &
       &
       &
       &
       &
       &
       &
       \\
    \textbf{SFT} &
      31.1 &
      62.8 &
      85.2 &
      \underline{39.1} &
      53.3 &
      54.3 &
      76.2 &
      25.8 &
      45.7 &
      49.2 \\
    \textbf{SFT$_\text{KL}$} &
      13.0 &
      45.2 &
      70.2 &
      26.5 &
      36.3 &
      38.2 &
      33.3 &
      22.2 &
      30.4 &
      28.6 \\ \midrule
      {\color[HTML]{656565}\textbf{\textit{Reinforcement Learning}}} &
       &
       &
       &
       &
       &
       &
       &
       &
       &
       \\ 
       \textbf{RL$_\text{GRPO}$}~\citep{shao2024deepseekmath} &
      24.7 &
      61.6 &
      79.2 &
      33.7 &
      47.1 &
      49.3 &
      75.6 &
      31.3 &
      42.1 &
      49.7 \\
    \textbf{SimpleRL-Zero}$^*$~\citep{zeng2025simplerl} &
      27.0 &
      54.9 &
      76.0 &
      25.0 &
      34.7 &
      43.5 &
      30.2 &
      23.2 &
      34.5 &
      29.3 \\
    \textbf{OpenReasoner-Zero}$^*$~\citep{hu2025open} &
      16.5 &
      52.1 &
      82.4 &
      33.1 &
      47.1 &
      46.2 &
      66.2 &
      29.8 &
      \textbf{58.7} &
      51.6 \\
    \textbf{PRIME-Zero}$^*$~\citep{cui2025process} &
      17.0 &
      54.0 &
      81.4 &
      39.0 &
      40.3 &
      46.3 &
      73.3 &
      18.2 &
      32.7 &
      41.4 \\
    \textbf{Oat-Zero}$^*$~\citep{zeng2025simplerl} &
      33.4 &
      61.2 &
      78.0 &
      34.6 &
      43.4 &
      50.1 &
      70.1 &
      23.7 &
      41.7 &
      45.2 \\ \midrule
      {\color[HTML]{656565}\textbf{\textit{SFT and RL}}} &
       &
       &
       &
       &
       &
       &
       &
       &
       &
       \\ 
    \textbf{SFT $\rightarrow$ RL} &
      \underline{32.5} &
      67.1 &
      84.2 &
      34.1 &
      54.6 &
      54.5 &
      76.4 &
      37.9 &
      49.6 &
      54.6 \\
    \textbf{LUFFY}~\citep{yan2025learning} &
       29.4 &
       65.6 &
       \underline{87.6} &
       37.5 &
       \underline{57.2} &
      55.5 &
      80.5 &
      39.9 &
      53.0 &
      \underline{57.8} \\
    \textbf{TAPO}$^*$~\citep{wu2025thought} &
       33.3 &
       \textbf{77.5} &
       83.4 &
       38.2 &
       46.2 &
      \underline{55.7} &
      \underline{81.6} &
      37.9 &
      49.6 &
      56.4 \\
    \textbf{ReLIFT}~\citep{ma2025learning} &
       28.2 &
       64.8 &
       85.0 &
       37.1 &
       54.9 &
      54.0 &
      74.9 &
      \underline{40.9} &
      51.9 &
      55.9 
       \\ \hdashline
       \rowcolor[HTML]{F0F0FF}\textbf{\method (ours)} &
       \textbf{35.3} &
       \underline{74.3} &
       \textbf{89.8} &
       \textbf{39.7} &
       \textbf{58.3} &
      \textbf{59.5} &
      \textbf{85.3} &
      \textbf{46.4} &
      \underline{55.9} &
      \textbf{62.5} \\ \bottomrule
    \end{tabular}%
    }
    \vspace{0.2em}
    \raggedleft{\scriptsize $^*$This method's performance is taken from the corresponding paper.}
    \end{table}
}

\subsection{Experimental Results}

\paragraph{Reasoning Benchmark Performance.}
Our main results are shown in Table~\ref{tab:main_performance}, where we compare \method with several zero-RL baselines, as well as direct SFT, and SFT+RL methods.
Across five challenging competition-level reasoning benchmarks, \method achieves an average score of \textbf{59.1}, significantly outperforming existing RL methods by a margin of \textbf{+9.0} points over the best baseline, clearly demonstrating the benefit of integrating demonstrations with self-exploration in LLM reasoning.
We also observe that \method achieves a \textbf{+4.8} points improvement over the SFT methods, indicating that the self-exploration component can effectively refine the policy distribution learned from demonstrations.
Compared to the SFT+RL methods, \method achieves a \textbf{+3.4} points improvement, demonstrating that the single-stage design and entropy-aware weighting mechanism can effectively balance the benefits of demonstrations and self-exploration.

\begin{figure}[t]
    \centering
	\subfigure[Training Rewards]{
    \begin{minipage}[t]{0.315\textwidth}
        \centering
        \includegraphics[width=\textwidth]{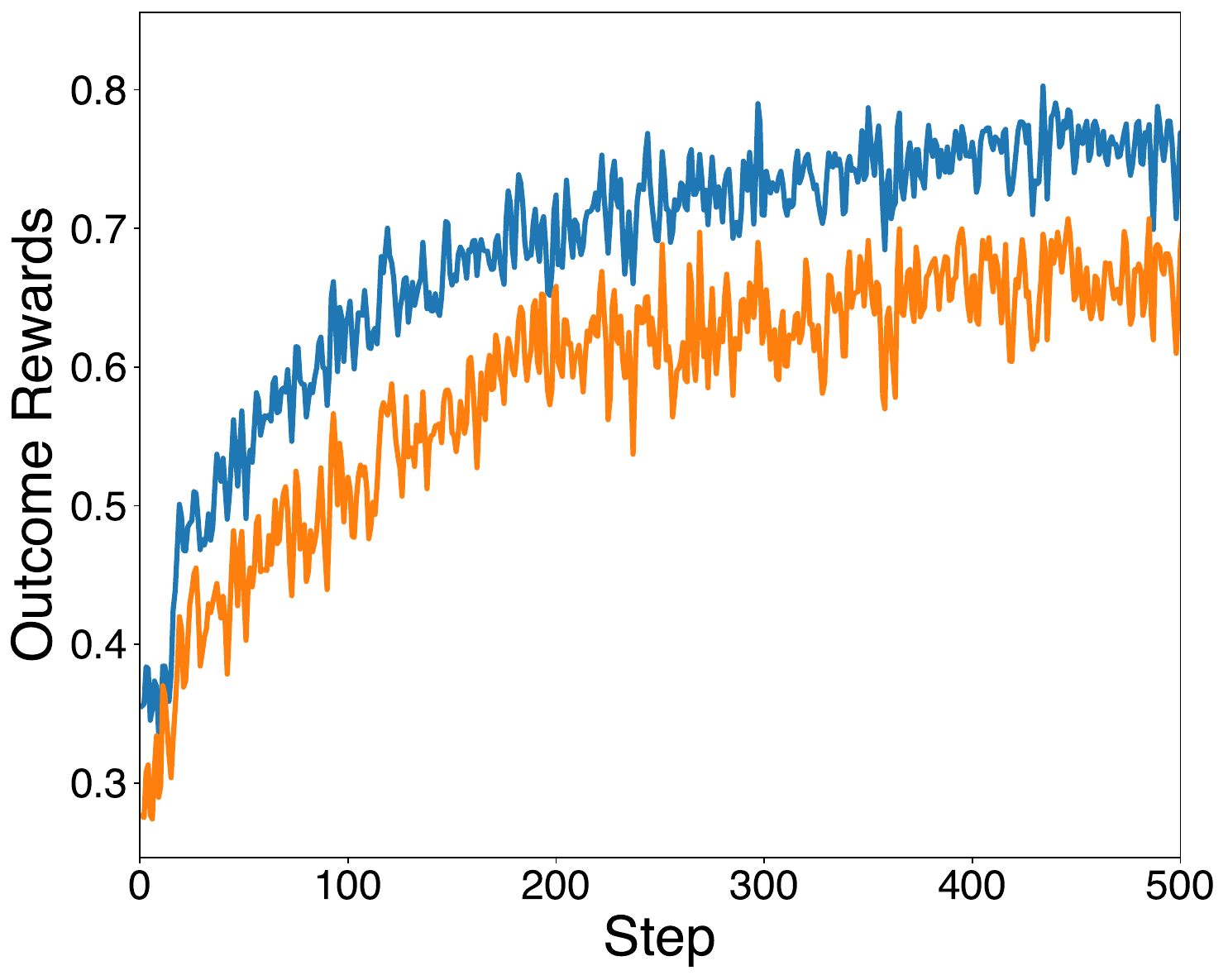}
        \label{fig:reward_comparison}
    \end{minipage}
	}
	\subfigure[Response Lengths]{
    \hfill
    \begin{minipage}[t]{0.315\textwidth}
        \centering
        \includegraphics[width=\textwidth]{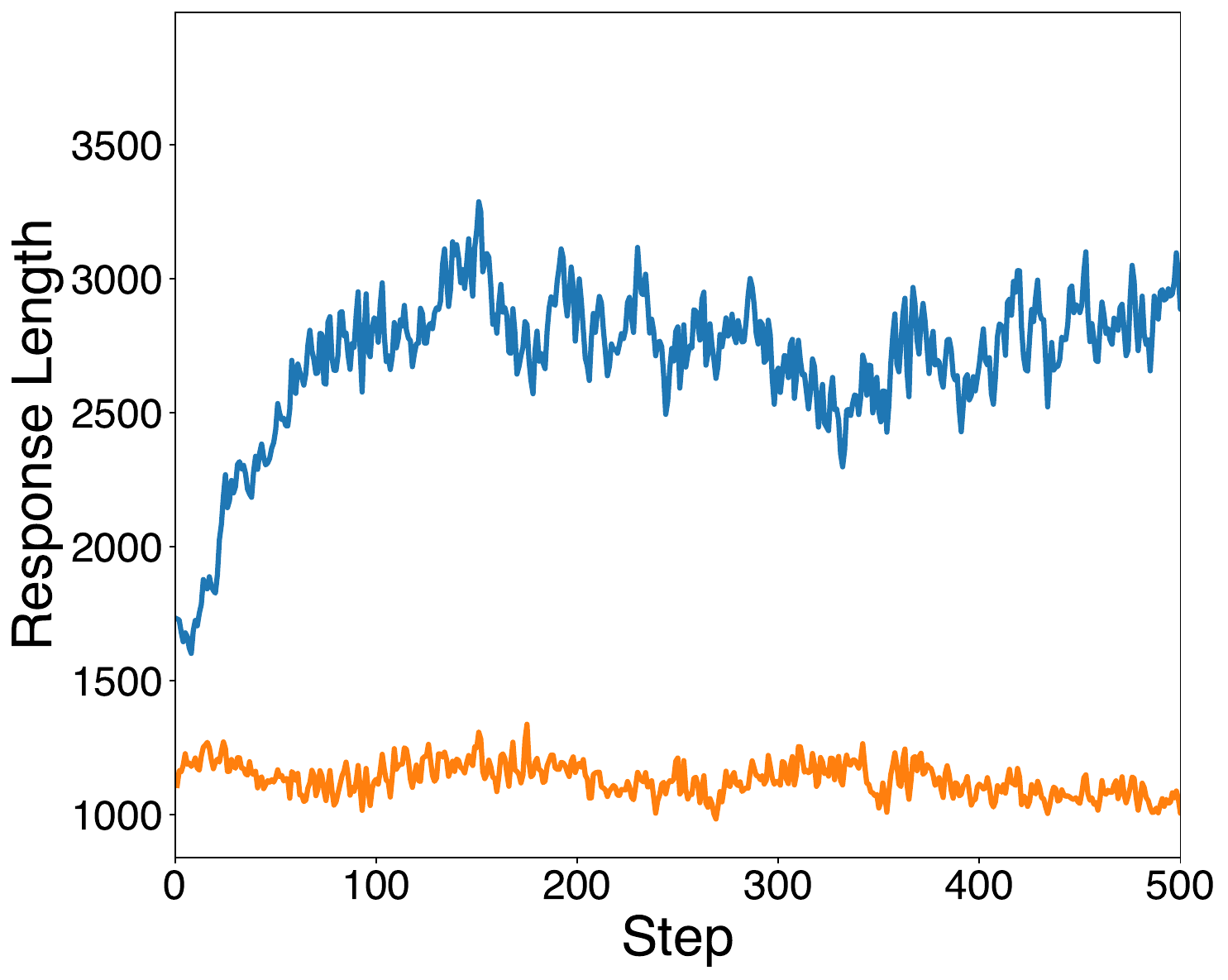}
        \label{fig:lengths_comparison}
    \end{minipage}
	}
	\subfigure[Training Entropy]{
    \hfill
    \begin{minipage}[t]{0.315 \textwidth}
        \centering
        \includegraphics[width=\textwidth]{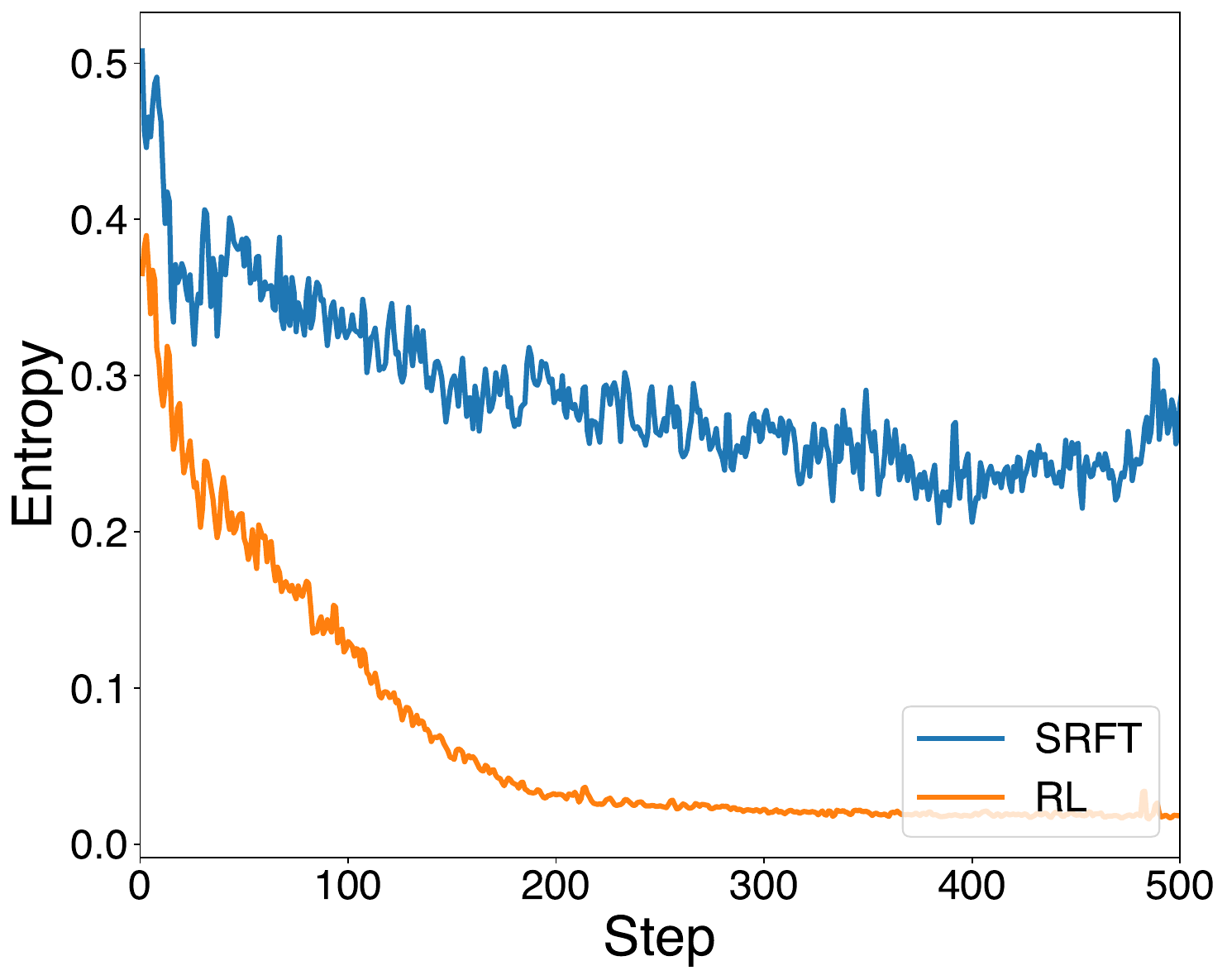}
        \label{fig:entropy_comparison}
    \end{minipage}
	}
    \caption{Training dynamics during RL and \method training, including training rewards, response lengths, and training entropy.}
    \label{fig:learning_dynamics}
\end{figure}

\paragraph{Out-of-Distribution Generalization.}
Regarding out-of-distribution performance, the results in Table~\ref{tab:main_performance} show that \method also achieves an average score of \textbf{62.5} and outperforms the best baseline by \textbf{+4.7} points. These results highlight \method's effectiveness in combining demonstrations with self-exploration to improve generalization ability.

\paragraph{Training Dynamics.}
Figure~\ref{fig:learning_dynamics} shows the training dynamics of \method, including training rewards, response lengths, and training entropy.
As shown in Figure~\ref{fig:reward_comparison}, \method achieves faster performance improvement compared to RL, with both \method and RL exhibiting an increasing trend in training rewards.
In terms of response length, as shown in Figure~\ref{fig:lengths_comparison}, when faced with challenging training data, RL exhibits a tendency toward generating more concise responses, whereas \method shows a progressive lengthening of responses, indicating the development of more thorough and detailed reasoning processes.
From an entropy perspective in Figure~\ref{fig:entropy_comparison}, compared to the rapid entropy decline exhibited by RL, our method \method maintains more stable entropy, indicating that the policy can continue exploring during training, which also demonstrates the effectiveness of the entropy-aware weighting mechanism.

\paragraph{Ablation Study.}
We conduct an ablation study to assess the effectiveness of each component.
As shown in Table~\ref{tab:ablations}, we evaluate the impact of the two key entropy-aware weighting mechanisms: $w_\text{SFT}$ for demonstrations learning and $w_\text{RL}$ for positive self-exploration samples. 
We ablate these mechanisms by setting their values to a fixed constant of $1.0$ to evaluate the contribution of these two weighting components.
Removing the SFT weighting mechanism (w/o $w_\text{SFT}$) results in a performance drop of \textbf{-4.0} points, while removing the RL weighting (w/o $w_\text{RL}$) leads to a \textbf{-2.9} point decrease, demonstrating that both components contribute significantly to overall performance.
The ablation results validate our theoretical analysis, confirming that entropy-aware weighting mechanisms enable \method to dynamically balance supervised learning and reinforcement learning components, leading to more stable training and superior performance compared to fixed weighting schemes.

{\addtolength{\extrarowheight}{\belowrulesep}
\aboverulesep=0pt
\belowrulesep=0pt
\begin{table}[t]
    \centering
    \renewcommand{\arraystretch}{1.1}
    \caption{Ablation results on \method, including the impact of $w_\text{SFT}$ and $w_\text{RL}$.}
    \label{tab:ablations}
    \resizebox{0.7\textwidth}{!}{%
    \begin{tabular}{>{}lccccc|c<{}}
    \toprule
    \textbf{Model} & \textbf{AIME24} & \textbf{AMC} & \textbf{MATH-500} & \textbf{Minerva} & \textbf{Olympiad} & \textbf{Avg.} \\ \midrule
    \textbf{Qwen2.5-Math}               & 11.4 & 32.6 & 48.8 & 8.7  & 15.8 & 23.5 \\
    \textbf{\method} w/o $w_\text{SFT}$ & 30.1 & 65.8 & 87.0 & 36.8 & 55.8 & 55.1 \\
    \textbf{\method} w/o $w_\text{RL}$ & 32.6 & 67.2 & 87.5 & 37.4 & 56.5 & 56.2 \\ \hdashline
    \rowcolor[HTML]{F0F0FF}
    \textbf{\method}                    & \textbf{35.3} & \textbf{72.2} & \textbf{89.8} & \textbf{39.7} & \textbf{58.3} & \textbf{59.1} \\ \bottomrule
    \end{tabular}%
    }
    \end{table}
}

\section{Related Work}

\paragraph{Reinforcement Learning for LLM Reasoning.}
The pursuit of complex reasoning capabilities in LLMs has witnessed remarkable progress, with RL emerging as a pivotal methodology for enhancing reasoning abilities beyond the limitations of SFT alone. Recent approaches such as GRPO~\citep{shao2024deepseekmath,guo2025deepseek}, DAPO~\citep{yu2025dapo}, DR.GRPO~\citep{liu2025understanding}, and VAPO~\citep{yue2025vapo} have demonstrated substantial improvements in mathematical reasoning and complex problem-solving tasks.
However, the precise mechanisms through which RL enhances reasoning capabilities remain incompletely understood. Several empirical investigations suggest that reinforcement learning primarily serves to elicit, refine, or improve the sampling of pre-existing reasoning abilities rather than instilling entirely novel fundamental reasoning skills from scratch. For instance, \citet{yue2025does} question whether current reinforcement learning with verifiable rewards (RLVR) genuinely expands the reasoning boundary (pass@k) or primarily improves the sampling efficiency of already known solutions (pass@1). Similarly, \citet{wang2025reinforcement} highlight that base models already possess substantial reasoning capabilities that reinforcement learning can effectively unlock or redirect. Nevertheless, ProRL~\citep{liu2025prorl} demonstrates that RL-trained models can achieve improved success rates on tasks where base models completely fail, suggesting that sustained and stable reinforcement learning training can indeed expand the reasoning capability boundaries of LLMs.
In this work, we design a single-stage method that combines SFT and RL, maintaining stable entropy during training and achieving continuous performance improvement.

\paragraph{Integrating Supervised Fine-Tuning and Reinforcement Learning.}
The synergistic interaction between SFT and RL represents a critical area of investigation in modern LLM development. SFT on high-quality reasoning chains can establish a robust initial policy foundation, which RL can subsequently optimize. \citet{cai2025much} explore the necessary extent of exploration following SFT, finding that RL continues to provide substantial benefits by enabling models to deviate from potentially suboptimal SFT trajectories. Recent research suggests that SFT may equip models with structured reasoning templates that RL subsequently validates and improves~\citep{chen2025sft}. Nevertheless, determining the optimal strategy for combining these complementary paradigms remains an active area of debate.
To enhance sample efficiency and provide structured guidance for RL exploration, researchers have investigated various approaches for integrating external supervision into the reinforcement learning framework. UFT~\citep{liu2025uft} proposes a novel paradigm that merges SFT and RL into a single process, using informative supervision signals like hints from partial solutions to guide exploration and accelerate convergence. Addressing the limitations of on-policy learning, LUFFY~\citep{yan2025learning} augments RLVR by incorporating off-policy reasoning traces from stronger models, dynamically balancing imitation with on-policy exploration to improve capabilities. 
ReLIFT~\citep{ma2025learning} addresses the limitations of pure RL by interleaving reinforcement learning with supervised fine-tuning on high-quality demonstrations collected during training, enabling models to acquire new knowledge beyond their original capabilities.
TAPO~\citep{wu2025thought} enhances RL by incorporating external high-level guidance in the form of "thought patterns" abstracted from prior samples, adaptively integrating these to balance model-internal exploration with external strategy exploitation. SASR~\citep{chen2025step} offers a hybrid framework that theoretically unifies SFT and RL, using SFT for initial warm-up and then adaptively blending it with an online RL method based on training dynamics to maintain core reasoning while exploring diverse paths, using high-quality SFT demonstrations as a key external data source. 
Furthermore, the single-stage integration of SFT and RL helps mitigate the catastrophic forgetting problem that previous methods encountered when transitioning from SFT to RL~\citep{chen2025step,liu2025superrl}.
These approaches collectively underscore an emerging trend toward more sophisticated integrations of supervised signals within reinforcement learning frameworks to improve reasoning alignment and overall performance.

\section{Conclusion}

In this work, we investigate the integration of SFT and RL for LLM reasoning. Through comprehensive analysis, we reveal that SFT performs coarse-grained global adjustments while RL conducts fine-grained selective optimizations, with entropy serving as a crucial training indicator. Building on these observations, we propose \method, a single-stage approach that unifies both paradigms through entropy-aware weighting mechanisms. Extensive experiments demonstrate \method's effectiveness, achieving 59.1\% average accuracy and outperforming zero-RL baselines by 9.0\% on reasoning tasks and 10.9\% on out-of-distribution benchmarks.

\textbf{Limitations.} 
While our work demonstrates the effectiveness of entropy-aware SFT-RL single-stage integration, our current utilization of entropy dynamics remains relatively simple with basic exponential weighting functions. The rich temporal patterns of entropy during training suggest opportunities for more sophisticated entropy-based control mechanisms. Future work could explore adaptive entropy scheduling or multi-timescale entropy analysis to better capture the interplay between SFT and RL signals, potentially leading to more principled hybrid training algorithms. Additionally, our approach assumes access to high-quality demonstrations, and future research could investigate the potential for training with imperfect demonstrations to enhance the method's applicability.

\newpage
\bibliographystyle{unsrtnat}
\bibliography{references}

\newpage
\appendix

\renewcommand{\thefigure}{A\arabic{figure}}
\renewcommand{\thetable}{A\arabic{table}}

\setcounter{figure}{0}
\setcounter{table}{0}

\renewcommand{\theequation}{A\arabic{equation}}
\setcounter{equation}{0}

\textbf{\LARGE{Appendix}}

\vspace{5pt}

\startcontents[section]
\printcontents[section]{l}{1}{\setcounter{tocdepth}{2}}

\newpage

\section{More Experimental Results}
\label{sec:more_experiments}

\subsection{Token Probability Visualization of \method}
\begin{figure}[ht]
	\centering
	\includegraphics[width=0.8\textwidth]{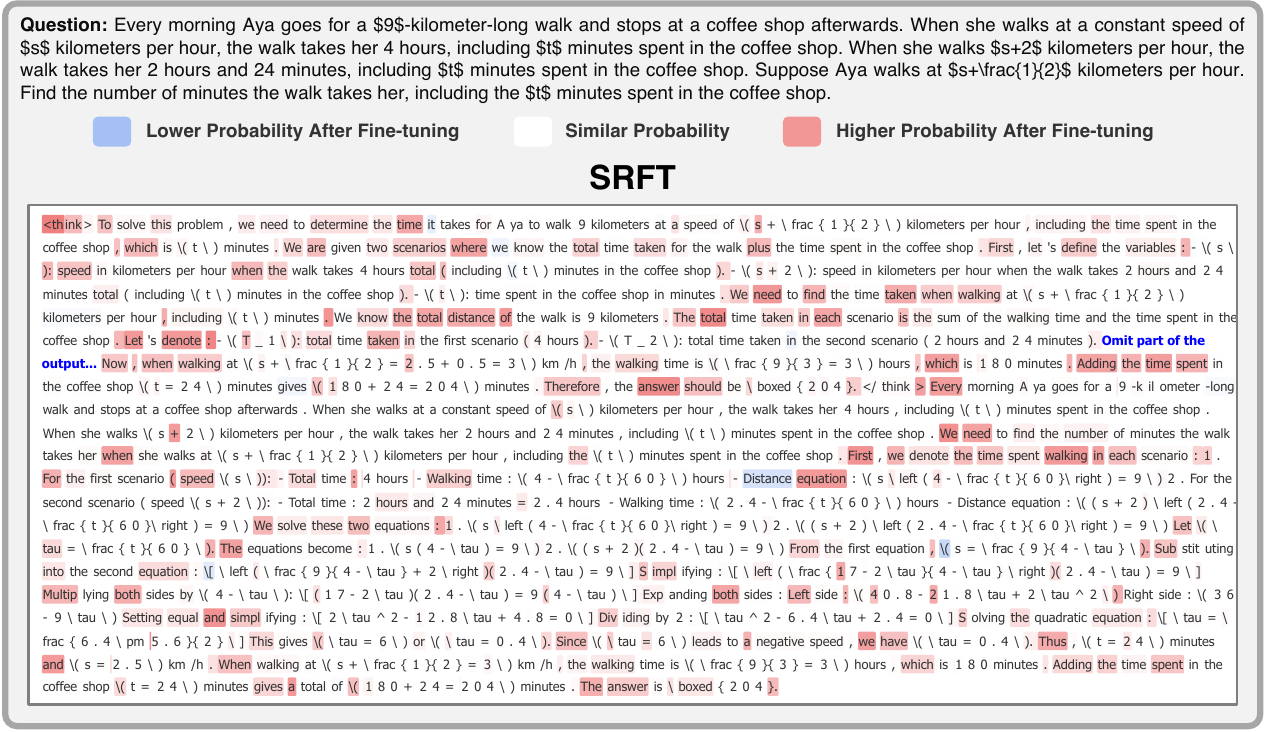}
	\caption{Token probability distribution visualization for \method.}
	\label{fig:token_prob_ours}
\end{figure}

We visualize the token probability distribution of \method after training, as shown in Figure~\ref{fig:token_prob_ours}.
We observe that the token probability changes are moderate, achieving a balanced point between SFT and RL that enhances the model's reasoning capabilities while preserving its base abilities.

\subsection{Entropy-aware Gradient Clipping}
\begin{wrapfigure}[17]{r}{0.46\textwidth}
	\centering
	\includegraphics[width=0.46\textwidth]{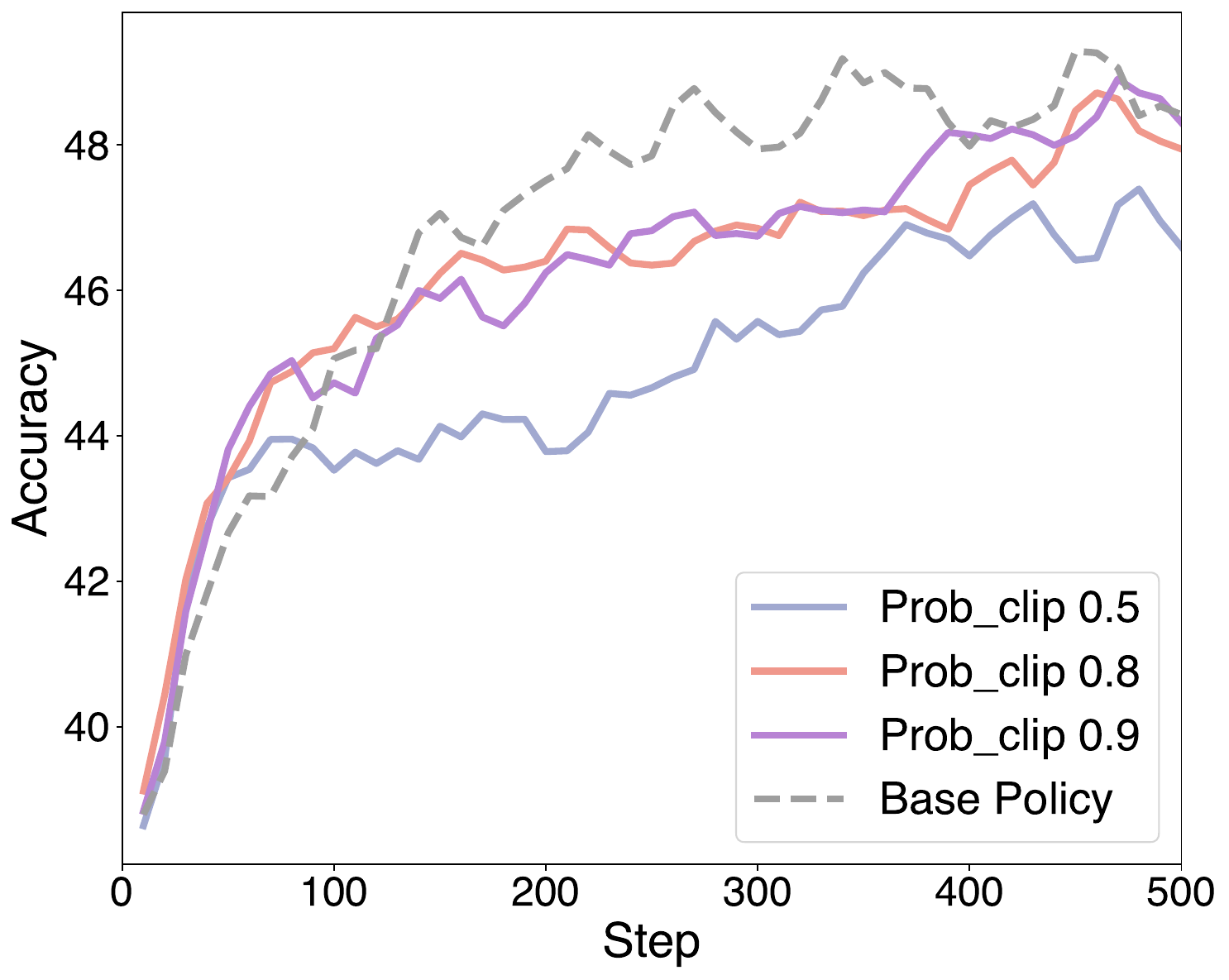}
	\vspace{-2em}
	\caption{Performance of RL with gradient clipping for low-entropy tokens.}
	\label{fig:prob_clip}
\end{wrapfigure}
Our investigation into the entropy characters of tokens modified by fine-tuning reveals that reinforcement learning predominantly targets tokens with high entropy distributions, a finding that aligns with recent work on selective optimization in language models~\citep{wang2025beyond}.
To empirically validate this observation, we design controlled experiments implementing gradient truncation for high-probability tokens during RL training. As demonstrated in Figure~\ref{fig:prob_clip}, the model's performance remains comparable to the original RL algorithm even when gradients are truncated for low-entropy tokens, providing strong empirical support for our hypothesis. This evidence confirms that RL operates with remarkable selectivity, precisely adjusting tokens with uncertain distributions while leaving confident predictions largely unchanged. In contrast, SFT applies broad modifications across the entire token space, fundamentally altering the model's distributional characters in a less discriminative manner.

\newpage
\section{Experimental Details}
\label{sec:experimental_details}

\textbf{Training.} We follow the SFT configuration of OpenR1-Qwen-7B~\citep{openr1}, performing full fine-tuning on DeepSeek-R1 generated reasoning traces and prompts. The training hyperparameters include a batch size of 128, learning rate of $5 \times 10^{-6}$, linear learning rate schedule with 10\% warmup, and training for 3 epochs. For SFT with KL regularization, we use identical settings while adding a KL divergence regularization between the current policy and the base model (Qwen2.5-Math-7B) with weight $\lambda = 0.2$. The SFT$_\text{KL}$ loss is:
\begin{equation}
\mathcal{L}_{\text{SFT}_{\text{KL}}}(\theta) = \mathbb{E}_{(\bm{x}, \bm{y}) \sim \mathcal{D}}[-\log \pi_{\theta}(\bm{y}|\bm{x})] + \lambda \mathcal{L}_{\text{KL}}(\theta,\theta_{\text{base}}),
\end{equation}
where $\mathcal{L}_{\text{KL}}(\theta,\theta_{\text{base}})$ is the KL divergence between the current policy and the base model.
For RL, we train for 500 steps with 8 rollouts per prompt. The learning rate is fixed at $1 \times 10^{-6}$.
For our method \method, we use the same training settings as RL.
Since the maximum sequence length for Qwen2.5-Math-7B is 4096, which is insufficient for our tasks, we increase the RoPE theta from 10,000 to 40,000 and expand the window size to 16,384.
For all experiments, we use verl\footnote{\url{https://github.com/volcengine/verl}}~\citep{sheng2024hybridflow} as the implementation framework. All experiments are conducted on 64 A100 GPUs.

\textbf{Evaluation.} All evaluations are conducted using VLLM~\citep{kwon2023efficient} with temperature set to 0.6 and maximum generation length of 8192 tokens. 
For datasets with limited sample sizes (AIME24 and AMC), we report the avg@32 metric; for the remaining three datasets, we adopt pass@1 as the evaluation criterion. 
We verify the correctness of generated solutions using Math-Verify and OAT-Grader~\citep{liu2024oat}.
For baseline comparisons, we independently validate the results of base model, SFT-related baselines, GRPO~\citep{shao2024deepseekmath}, LUFFY~\citep{yan2025learning}, and ReLIFT~\citep{ma2025learning}, while results for TAPO~\citep{wu2025thought} and other zero-shot reinforcement learning models are taken from the TAPO (because we cannot find the open-source code or model) and LUFFY papers. 

\textbf{Reward Design.} To evaluate the impact of our method, we adopt a simple reward function as
below. All training experiments employ the same reward function.
\begin{equation}
R(\bm{x}, \bm{y}) = 
\begin{cases}
	1, & \text{if } \bm{y} \text{ is correct} \\
0, & \text{otherwise}
\end{cases}.
\end{equation}

\begin{figure}[ht]
	\centering
	\includegraphics[width=0.8\textwidth]{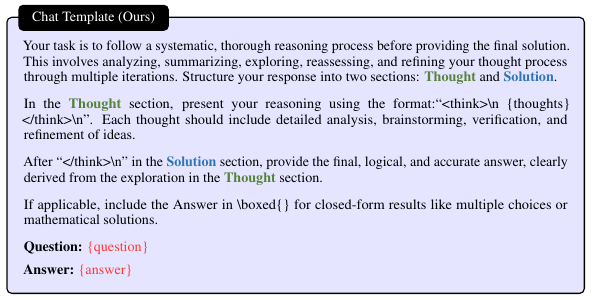}
	\caption{Chat template for all training paradigms (SFT, RL, \method).}
	\label{fig:chat_template}
\end{figure}

\begin{figure}[ht]
	\centering
	\includegraphics[width=0.8\textwidth]{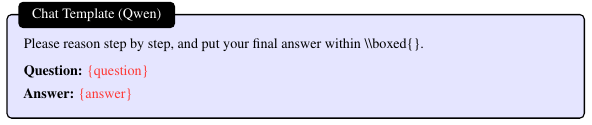}
	\caption{Chat template for Qwen-2.5-Math.}
	\label{fig:qwen_ct}
\end{figure}

\textbf{Chat Template.}
Following \citet{yan2025learning,ma2025learning}, for all training paradigms (SFT, RL, \method), we employ a unified system prompt that encourages systematic reasoning, as shown in Figure~\ref{fig:chat_template}.
We also experimented with alternative templates, as shown in Figure~\ref{fig:qwen_ct}.

\textbf{Template Ablation.} To minimize template influence, we evaluated the base Qwen-7B-Math model with different templates. Results are shown in Table~\ref{tab:template_ablation}, which indicates that our template design effectively guides the model's reasoning process while maintaining consistency across different mathematical domains.

\begin{table}[ht]
	\centering
	\renewcommand{\arraystretch}{1.2}
	\caption{Template ablation results on mathematical reasoning benchmarks}
	\label{tab:template_ablation}
	\begin{tabular}{lcccccc}
	\toprule
	\textbf{Template} & \textbf{AIME24} & \textbf{AMC} & \textbf{MATH500} & \textbf{Minerva} & \textbf{Olympiad} & \textbf{Average} \\
	\midrule
	\textbf{No Template} & 0.302 & 0.132 & 0.596 & 0.424 & 0.134 & 0.318 \\
	\textbf{Qwen2.5-Math} & 0.144 & 0.088 & 0.446 & 0.303 & 0.111 & 0.218 \\
	\textbf{\method} & 0.296 & 0.165 & 0.648 & 0.448 & 0.141 & 0.340 \\
	\bottomrule
	\end{tabular}
	\end{table}

\section{Dataset Details}
\label{sec:dataset_details}

\subsection{Training Dataset}
\textbf{OpenR1-Math-220k} is a comprehensive dataset designed for mathematical reasoning, comprising 220,000 math problems. Each problem is associated with two to four reasoning traces generated by the DeepSeek-R1 model. The traces have been verified using tools like Math Verify and Llama-3.3-70B-Instruct, ensuring at least one correct reasoning path per problem. This dataset challenges models to understand and replicate complex reasoning processes across various mathematical domains including algebra, geometry, number theory, and calculus.

\subsection{Evaluation Benchmarks}

To evaluate the models above, we use eight benchmarks categorized into mathematical reasoning benchmarks and out-of-distribution (OOD) benchmarks as described below.

\subsubsection{Mathematical Reasoning Benchmarks}

{\addtolength{\extrarowheight}{\belowrulesep}
\aboverulesep=0pt
\belowrulesep=0pt
\begin{table}[t]
    \centering
    \renewcommand{\arraystretch}{1.1}
    \caption{Benchmarks used in this study. ``--'' indicates the split is not officially provided.}
    \label{tab:datasets}
    \resizebox{\textwidth}{!}{%
    \begin{tabular}{ccccccc}
    \toprule
    \textbf{Dataset} & \textbf{\#Train} & \textbf{\#Test} & \textbf{Task Type} & \textbf{Domain} & \textbf{License} & \textbf{Source} \\
    \midrule
    \multicolumn{7}{c}{\textbf{Training Dataset}} \\
    \midrule
    \textsc{OpenR1-Math-220k} & 220,000 & -- & Math reasoning & Mathematics & Apache 2.0 & \href{https://huggingface.co/datasets/open-r1/OpenR1-Math-220k}{[Link]} \\
    \midrule
    \multicolumn{7}{c}{\textbf{Mathematical Reasoning Benchmarks}} \\
    \midrule
    \textsc{AIME24} & -- & 30 & Math competition & Mathematics & MIT & \href{https://huggingface.co/datasets/Maxwell-Jia/AIME_2024}{[Link]} \\
    \textsc{AMC} & -- & 83 & Math competition & Mathematics & Apache 2.0 & \href{https://huggingface.co/datasets/AI-MO/aimo-validation-amc}{[Link]} \\
    \textsc{MATH500} & -- & 500 & Mathematical reasoning & Mathematics & - & \href{https://huggingface.co/datasets/HuggingFaceH4/MATH-500}{[Link]} \\
    \textsc{Minerva} & -- & 272 & Mathematical reasoning & Mathematics & Apache 2.0 & \href{https://huggingface.co/datasets/math-ai/minervamath}{[Link]} \\
    \textsc{Olympiad} & -- & 674 & Math competition & Mathematics & Apache 2.0 & \href{https://huggingface.co/datasets/Hothan/OlympiadBench/viewer/OE_TO_maths_en_COMP?views\%5B\%5D=oe_to_maths_en_comp}{[Link]} \\
    \midrule
    \multicolumn{7}{c}{\textbf{Out-of-Distribution (OOD) Benchmarks}} \\
    \midrule
    \textsc{ARC-C} & -- & 1{,}172 & Science reasoning & General science & CC-BY-SA-4.0 & \href{https://huggingface.co/datasets/allenai/ai2_arc}{[Link]} \\
    \textsc{GPQA-D} & -- & 198 & Scientific reasoning & Bio, Phys, Chem & CC-BY-4.0 & \href{https://huggingface.co/datasets/Idavidrein/gpqa}{[Link]} \\
    \textsc{MMLU-Pro} & -- & 12{,}032 & Multi-task understanding & Multidisciplinary & MIT & \href{https://huggingface.co/datasets/TIGER-Lab/MMLU-Pro}{[Link]} \\
    \bottomrule
    \end{tabular}
    }
    \end{table}
}

\textbf{AIME24} is a benchmark dataset based on problems from the 2024 American Invitational Mathematics Examination, a prestigious high school mathematics competition in the United States. The AIME24 benchmark tests a model's ability to solve challenging mathematics problems by generating step-by-step solutions and providing the correct answers. This dataset contains problems from the American Invitational Mathematics Examination (AIME) 2024, organized in JSONL format where each line represents a complete problem. Concepts typically covered include topics in elementary algebra, geometry, trigonometry, as well as number theory, probability, and combinatorics. The examination consists of 15 problems with integer answers between 0 and 999, requiring advanced mathematical reasoning and problem-solving skills.

\textbf{AMC} is a validation dataset containing problems from the American Mathematics Competitions, specifically AMC12 from 2022 and 2023. All 83 problems come from AMC12 2022, AMC12 2023, and have been extracted from the AOPS wiki page. The AMC 10 is a 25-question, 75-minute multiple-choice competition designed for students in grades 10 and below. The content covers mathematical topics such as elementary algebra, basic geometry, area and volume formulas, elementary number theory, and elementary probability. This dataset serves as an internal validation set and focuses on competition-level mathematical problems comparable in difficulty to AMC12 and AIME exams.

\textbf{MATH500} is a carefully curated subset of mathematical problems designed for robust evaluation. MATH500 is a subset of 500 randomly sampled questions from Hendrycks' 2021 MATH dataset, created by OpenAI in late 2024 as a consequence of their appropriation of 90\% of the original 5000 MATH questions for training data for reinforcement learning on o1-series models. The dataset maintains the diversity and complexity of the original MATH benchmark while providing a clean evaluation set that avoids potential data contamination issues.

\textbf{Minerva} is a mathematical reasoning benchmark that encompasses a wide range of mathematical domains and difficulty levels. The dataset is designed to evaluate models' capabilities in advanced mathematical problem-solving, including topics from high school to undergraduate level mathematics. It includes problems requiring multi-step reasoning, symbolic manipulation, and deep mathematical understanding across various mathematical fields.

\textbf{Olympiad} refers to mathematical olympiad-level problems that represent some of the most challenging mathematical reasoning tasks. Unlike existing Olympiad-related benchmarks, datasets in this category focus exclusively on mathematics and comprise vast collections of competition-level problems. These problems are meticulously categorized into 33+ sub-domains and span across 10+ distinct difficulty levels. These problems require exceptional mathematical insight, creativity, and advanced problem-solving techniques typically seen in international mathematical competitions.  We specifically utilize the \textsc{OE\_TO\_maths\_en\_COMP} subset for our evaluation.

\subsubsection{Out-of-Distribution (OOD) Benchmarks}

\textbf{ARC-C} (AI2 Reasoning Challenge-Challenge) is a dataset of grade-school level science questions that require commonsense reasoning and knowledge application. The dataset consists of multiple-choice questions that are designed to be easy for humans but challenging for AI systems, testing the model's ability to apply scientific knowledge and reasoning in everyday contexts beyond pure mathematical domains.

\textbf{GPQA-D} (Graduate-Level Google-Proof Q\&A-Diamond) is a challenging benchmark designed to evaluate advanced reasoning capabilities in scientific domains. GPQA consists of 448 multiple-choice questions designed to evaluate the capabilities of LLMs and scalable oversight mechanisms. This dataset provides "Google-proof" questions in biology, physics, and chemistry, designed to test deep domain expertise and reasoning under challenging conditions. The diamond subset contains 198 hard problems. The questions require graduate-level knowledge and are specifically designed to be difficult to answer even with internet search.

\textbf{MMLU-Pro} (Massive Multitask Language Understanding-Professional) is an enhanced version of the original MMLU benchmark designed to be more challenging and robust. The MMLU-Pro dataset is an enhanced version of the Massive Multitask Language Understanding benchmark. It's designed to be more robust and challenging, aiming to rigorously evaluate language understanding capabilities. MMLU is a comprehensive benchmark that covers 57 subjects across fields like mathematics, history, law, and medicine. It assesses not only factual knowledge but also the model's capacity to apply this knowledge in context-specific scenarios. MMLU-Pro increases the difficulty and reduces potential shortcuts while maintaining broad coverage across academic disciplines.

\newpage
\section{SFT Gradient Derivation}
\label{sec:sft_gradient_derivation}

In this section, we provide the detailed mathematical derivation for the gradient of the Supervised Fine-Tuning (SFT) loss function, which was referenced in Sec.~\ref{sec:token_distributions}.

\subsection{Problem Setup}

Given a dataset $\mathcal{D} = \{(\bm{x}_i, \bm{y}_i)\}_{i=1}^N$ where $\bm{x}_i$ is an input prompt and $\bm{y}_i = (y_{i,1}, y_{i,2}, \ldots, y_{i,T_i})$ is the corresponding target sequence, the SFT objective function is (the same as Eq.~\eqref{eq:sft_loss}):
\begin{equation}
\mathcal{L}_{\text{SFT}}(\theta) = \mathbb{E}_{(\bm{x}, \bm{y}) \sim \mathcal{D}} [-\log \pi_{\theta}(\bm{y} | \bm{x})].
\end{equation}
Since the sequence probability is factorized as:
\begin{equation}
\pi_{\theta}(\bm{y} | \bm{x}) = \prod_{t=1}^{T} \pi_{\theta}(y_t | \bm{x}, \bm{y}_{<t}).
\end{equation}
The SFT loss becomes:
\begin{equation}
\mathcal{L}_{\text{SFT}}(\theta) = \mathbb{E}_{(\bm{x}, \bm{y}) \sim \mathcal{D}} \left[ -\sum_{t=1}^{T} \log \pi_{\theta}(y_t | \bm{x}, \bm{y}_{<t}) \right].
\end{equation}

\subsection{Gradient Derivation}

To derive the gradient $\nabla_\theta \mathcal{L}_{\text{SFT}}(\theta)$, we need to compute:
\begin{equation}
\nabla_\theta \mathcal{L}_{\text{SFT}}(\theta) = \mathbb{E}_{(\bm{x}, \bm{y}) \sim D_{\text{data}}} \left[ -\sum_{t=1}^{T} \nabla_\theta \log \pi_{\theta}(y_t | \bm{x}, \bm{y}_{<t}) \right].
\end{equation}
Now, let's consider the key insight: at each time step $t$, the model produces a probability distribution over the entire vocabulary $\mathcal{V}$. The gradient of the log probability for the target token $y_t$ can be expressed in terms of the model's prediction probabilities for all vocabulary tokens.

For any token $v \in \mathcal{V}$ at position $t$, we have:
\begin{equation}
\nabla_\theta \log \pi_{\theta}(v | \bm{x}, \bm{y}_{<t}) = \frac{1}{\pi_{\theta}(v | \bm{x}, \bm{y}_{<t})} \nabla_\theta \pi_{\theta}(v | \bm{x}, \bm{y}_{<t})
\end{equation}
Since $\sum_{v \in \mathcal{V}} \pi_{\theta}(v | \bm{x}, \bm{y}_{<t}) = 1$, we have the constraint:
\begin{equation}
\sum_{v \in \mathcal{V}} \nabla_\theta \pi_{\theta}(v | \bm{x}, \bm{y}_{<t}) = 0.
\end{equation}
Using the chain rule and the fact that the softmax normalization affects all vocabulary tokens, the gradient can be written as:
\begin{equation}
\nabla_\theta \log \pi_{\theta}(y_t | \bm{x}, \bm{y}_{<t}) = \sum_{v \in \mathcal{V}} \left( \mathbf{1}_{v = y_t} - \pi_{\theta}(v | \bm{x}, \bm{y}_{<t}) \right) \nabla_\theta \log \pi_{\theta}(v | \bm{x}, \bm{y}_{<t}),
\end{equation}
where $\mathbf{1}_{v = y_t}$ is the indicator function that equals 1 when $v = y_t$ and 0 otherwise.

Substituting this back into the SFT gradient, we obtain:
\begin{equation}
\nabla_\theta \mathcal{L}_{\text{SFT}} = \mathbb{E}_{(\bm{x}, \bm{y}) \sim \mathcal{D}} \left[ \sum_{t=1}^{|\bm{y}|} \sum_{v \in \mathcal{V}} \left( \pi_\theta(v|\bm{x}, \bm{y}_{<t}) - \mathbf{1}_{v = y_t} \right) \nabla_\theta \log \pi_\theta(v|\bm{x}, \bm{y}_{<t}) \right].
\end{equation}
This formulation reveals the fundamental mechanism of SFT: at each time step, the gradient encourages the model to increase the probability of the target token ($\mathbf{1}_{v = y_t} = 1$) while decreasing the probabilities of all other tokens in the vocabulary ($\mathbf{1}_{v = y_t} = 0$). The magnitude of the decrease for each non-target token $v$ is proportional to its current probability $\pi_\theta(v|\bm{x}, \bm{y}_{<t})$.

This analysis confirms our empirical observations that SFT produces broad, coarse-grained changes to the model's probability distributions across the entire vocabulary, systematically sharpening the distribution toward the target tokens in the training data.

\section{Details of Visualization of Learning Dynamics}
\label{sec:model_trajectory}

To characterize model dynamics during training, in Sec.~\ref{sec:learning_dynamics}, we formulate a language model as a mapping from input prompts to output probability distributions over the vocabulary. Two models are considered similar or close if they assign identical probabilities to every output token across all input prompts. 
Under this equivalence, each model can be theoretically represented as a point in an infinite-dimensional probability space, where each dimension corresponds to the model's assigned probability for a specific token at a specific position within all possible outputs.

\subsection{Theoretical Definition}

\textbf{Model Space Formalization.}
We define the model space $\mathcal{M}$ as the set of all possible conditional probability distributions over token sequences. Each model $M \in \mathcal{M}$ can thus be represented as a vector in an infinite-dimensional space, with each dimension corresponding to the conditional probability $p_M(v_i \mid \bm{x}, \bm{y}_{<t})$ for vocabulary token $v_i$ at position $t$.

\textbf{Measure Model Dynamics via Reference Models.}
To make this space tractable for analysis and visualization, we measure the model dynamics by the distance between each model and a set of reference models $\mathcal{R} = \{R_1, R_2, \ldots, R_k\}$ (reference frames):
\begin{equation}
\bm{d}_{\mathcal{R}}(M) = \left(d_{R_1}(M), d_{R_2}(M), \ldots, d_{R_k}(M)\right)^\intercal,
\end{equation}
where $d_{R_i}(M)$ quantifies the distance between model $M$ and reference model $R_i$ in terms of assigned sequence probabilities.

\subsection{Experimental Setup}
\label{sec:dataset_generation}

To instantiate the above definition, we construct a dataset comprising 1,024 response sequences generated by each of the reference models under identical prompts. These prompts are drawn from a diverse mixture of mathematical reasoning benchmarks (including AIME24, Minerva, Olympiad, AMC, and MATH-500) to ensure broad coverage across problem domains and difficulty levels.

For each response, we compute the distance between the probabilities assigned by model $M$ to the reference tokens, and then aggregate overall responses to obtain the final distance $d_{R_i}(M)$ for each $R_i$. These distances collectively define the model's position in the projected subspace.
We select three reference models to construct the projection basis:
(1) DeepSeek-R1~\citep{guo2025deepseek}, representing state-of-the-art reasoning performance;
(2) QwQ-32B~\citep{qwq32b}, serving as a high-performing but structurally distinct baseline; and
(3) Qwen-2.5-Math-7B~\citep{yang2024qwen2}, which acts as the base model prior to fine-tuning.
Together, they span a spectrum from foundational to advanced capabilities, forming a semantically meaningful coordinate system.

For every model checkpoint throughout training, we evaluate its distance based on its probability assignments over the reference responses. This yields a trajectory in the model space that traces how the model evolves over time. By comparing these trajectories across training paradigms (e.g., SFT and RL), we uncover distinct optimization dynamics and convergence behaviors.
This three-dimensional distance framework provides both theoretical grounding and practical interpretability for analyzing training dynamics. It enables direct comparison of different model variants and reveals how specific training strategies influence progression through the reasoning capability landscape.

\end{document}